\documentclass[sigconf]{acmart}

\renewcommand\footnotetextcopyrightpermission[1]{} 
\usepackage{booktabs} 
\usepackage{xcolor}
\usepackage{colortbl}
\usepackage{bm}
\usepackage{tikz}
\usepackage{pgfplots}
\usepackage{tabularx}
\usepackage{multirow}
\usepackage{rotating}
\usepackage{xspace}
\usepackage[T1]{fontenc}

\usetikzlibrary{pgfplots.groupplots}
\pgfplotsset{every tick label/.append style={font=\small}}
\pgfplotsset{compat=1.13}
\setcopyright{rightsretained}

\newcommand{\ie}{i.\,e.\xspace}

\newcommand{\facts}{\mathcal{F}}
\newcommand{\claims}{\mathcal{C}}
\newcommand{\sources}{\mathcal{S}}
\newcommand{\tpr}{\mathit{tpr}}
\newcommand{\fpr}{\mathit{fpr}}

\renewcommand{\vec}{\bm}
\renewcommand{\paragraph}{\emph}

\newcommand{\dsCount}{10000}   

\colorlet{best}{black!20!white}
\colorlet{worst}{gray!80!white}
\colorlet{base}{red!70!white}
\newcommand{\colCol}[1]{%
\cellcolor{base>wheel,#1,360,2;gray,1}
}

\acmDOI{}
\acmISBN{}
\setcopyright{none}
\acmConference[]{}{}{} 

\normalsize

\begin{document}

\title{Restricted Boltzmann Machines for Robust and Fast Latent Truth Discovery}


\author{Klaus Broelemann}
\affiliation{%
  \institution{SCHUFA Holding AG, Innovation Lab}
  \city{Wiesbaden} 
  \country{Germany}
}
\email{Klaus.Broelemann@schufa.de}

\author{Thomas Gottron}
\affiliation{%
  \institution{SCHUFA Holding AG, Innovation Lab}
  \city{Wiesbaden} 
  \country{Germany}
}
\email{Thomas.Gottron@schufa.de}

\author{Gjergji Kasneci}
\affiliation{%
  \institution{SCHUFA Holding AG, Innovation Lab}
  \city{Wiesbaden} 
  \country{Germany}
}
\email{Gjergji.Kasneci@schufa.de}

\begin{abstract}
We address the problem of latent truth discovery, LTD for short, where the goal is to discover the underlying true values of entity attributes in the presence of noisy, conflicting or incomplete information. Despite a multitude of algorithms to address the LTD problem that can be found in literature, only little is known about their overall performance with respect to effectiveness (in terms of truth discovery capabilities), efficiency and robustness. A practical LTD approach should satisfy all these characteristics so that it can be applied to heterogeneous datasets of varying quality and degrees of cleanliness.  

We propose a novel algorithm for LTD that satisfies the above requirements. The proposed model is based on Restricted Boltzmann Machines, thus coined LTD-RBM. In extensive experiments on various heterogeneous and publicly available datasets, LTD-RBM is  superior to state-of-the-art LTD techniques in terms of an overall consideration of effectiveness, efficiency and robustness.
\end{abstract}


\maketitle

\section{Introduction}

Quality assessment for information coming from different sources is of tremendous importance in data integration tasks. 
Such tasks encompass information aggregation and information sharing applications, e-Commerce product descriptions, or even database consolidations after the merger of companies. 
For a huge number of Big Data applications information about the same objects can be obtained from different sources. 
These applications (such as booking or social editing and recommendation applications on the web) face the challenges of qualitative data integration.
In the absence of a ground truth, one of the main challenges when integrating information from multiple sources is contradicting information, which is typically the result of sources delivering noisy, outdated, erroneous, or incomplete information. 
For example, for the same flight, different flight booking web sites may report different arrival times. 
This challenge of deriving the most complete and accurate merged information for entities in such scenarios is commonly referred to as the \emph{latent truth discovery} (LTD) problem. 

State-of-the-art methods solving the LTD problem go well beyond casting a simple majority vote among all sources. Instead, they incorporate the reliability of the sources in their decisions and jointly infer source reliabilities and trustworthiness of information. Quite often the methods are based on certain assumptions about the underlying data and depend on parameters for a generative model of the observations.
The quality of a method is commonly measured in terms of  effectiveness (i.e. the quality of the discovered truths) and efficiency (concerning the runtime of the inference procedure).

One aspect that so far has not been addressed in this context is the \emph{robustness} of the LTD methods. 
By robustness we mean the effectiveness with respect to varying parameter settings, varying data quality and varying data set properties (such as underlying source accuracy, amount of copied claims or difficulty of the truth discovery process in terms of normalized entropy).
Hence, beyond the traditional quality criteria it is desirable for an LTD approach to show both: a stable behavior and high-quality predictions.

%

There are several proposals in literature to solve the LTD problem, e.g.,~\cite{yin2008truth,dong2009integrating,galland2010corroborating,kasneci2011cobayes,zhao2012bayesian,li2012truth,pasternack2013latent,li2014confidence,li2014resolving,wang2014using}; an overview can be found in~\cite{li2015survey}. However, a clean methodological comparison under all of the aspects of effectiveness, efficiency and robustness is difficult, mainly for the following reasons:

{\bf Heterogeneity and quality of the experimental datasets} The experimental datasets vary in structure and quality from article to article and are sometimes proprietary (e.g.~\cite{kasneci2010bayesian,kasneci2011cobayes}).

{\bf Nontransparent data preprocessing} Even when the datasets are freely available, the applied preprocessing steps are often nontransparent; it is important to note that the results are often highly dependent on the preprocessing steps. 
	
{\bf Data model diversity} Different methods build on different data models and model semantics; some consider binary values (e.g.~\cite{wang2012truth}), others multinomial (e.g.~\cite{zhao2012bayesian}) or even continuous values (e.g.~\cite{zhao2012probabilistic}).

{\bf Parameter dependency} All approaches come with their own parameters, which can be tuned to achieve better performance on specific experimental datasets; however, since in reality the ground truth would be missing, it would be desirable to have a robust setting for the used parameters that works well across various datasets.

As a practical solution should be a holistic approach that combines effectiveness with efficiency and robustness, these issues need to be addressed. With this goal in mind we provide the following contributions in this paper:
 
- We propose a novel LTD approach based on Restricted Boltzmann Machines (RBMs)~\cite{hinton2002training,fischer2012introductionRBM}. The underlying model is quite generic and can in principle handle discrete and continuous values.
 	
- We provide a practical inference procedure, based on \emph{Contrastive Divergence}~\cite{hinton2002training,carreira2005contrastive} and Gibbs sampling~\cite{casella1992explaining}, which is effective, efficient and robust in the above sense.

- In an extensive experimental evaluation on various heterogeneous datasets of varying quality levels we show that our solution, LTD-RBM, is superior to state-of-the-art LTD techniques in terms of an overall consideration of efficiency, quality and robustness.

This work extends our previous work\cite{Broelemann2017} by profound technical details, extensions, multiple proofs and experiments.
 The remainder of this paper is organized as follows: We give an overview of related work in Section~\ref{lab:related_work}. A general formalization of the LTD problem and the underlying data model are presented in ~\ref{sec:model}. Section~\ref{sec:algorithm} introduces the LTD model and the inference procedure. The experimental evaluation follows in Section~\ref{sec:experiments}, and we conclude in Section~\ref{sec:conclusion}.

\section{Related Work}
\label{lab:related_work}

From an abstract viewpoint, one possible way of categorizing latent truth discovery methods is by looking at the underlying inference and learning algorithms:

{\bf Bayesian Inference algorithms} use prior distributions for the truth and reliability parameters and jointly estimate truth and source reliability by fitting the parameters to the available data based on the assumed prior distributions.
	
{\bf Fix-point algorithms} start with an initial guess on the truth and reliability parameters and simplifying assumptions are used to iteratively fit the parameters to the available data.
	
{\bf Semi-supervised and Active Learning algorithms} start with a set of known ground truth labels. This initial ground truth and other assumptions are exploited to learn the reliability of sources. This reliability estimations can be used to estimate the latent truth. 

In the following paragraphs, we give an overview of the above three groups by highlighting representative approaches. 

\paragraph{Bayesian Inference}
AccuSim~\cite{dong2009integrating,li2012truth} integrates the similarity between claimed values into the Bayesian inference approach and proposes an extension of the algorithm AccuCopy in which source similarities -- in terms of which source might have copied from which other source -- are considered.
A Bayesian approach to knowledge corroboration is proposed  by~\cite{kasneci2011cobayes,kasneci2010bayesian}, where a latent truth discovery model integrates the logical dependencies between facts in a knowledge base and crowd opinions to derive the underlying correctness of the facts in the knowledge base.
Latent Truth Model (LTM)~\cite{zhao2012bayesian} is a probabilistic graphical model that applies collapsed Gibbs sampling to estimate the false positive and the false negative rate of sources by optimizing for the most probable answers.
Another Bayesian inference approach for continuous responses is presented in~\cite{zhao2012probabilistic}.

\paragraph{Fix-point Algorithms}
TruthFinder~\cite{yin2008truth} models the influence between claimed values and alternately estimates source reliabilities and the latent truth based on each other's values. 
In 2-Estimates~\cite{galland2010corroborating} the assumption that there is one and only one true value for each object is integrated in a voting-based  fix-point algorithm. The authors also propose an extension, 3-Estimates, in which the difficulty of deriving the true value of an object is considered.
In~\cite{pasternack2010knowing}, a source uniformly "invests" its reliability among the values it has claimed for the objects. The confidence of a value grows according to a non-linear function defined on the sum of invested reliabilities. In turn, the sources collect credits back from the confidence of their claimed values.
A Maximum Likelihood formulation of latent truth discovery for crowd/social sensing applications is provided by~\cite{wang2012truth,wang2013recursive}. The Expectation Maximization algorithm is proposed to derive the most probable answers as well as the true positive and true negative rates of sources (human agents in this case).

\paragraph{Semi-supervised and Active Learning Algorithms}
In~\cite{yin2011semi} a semi-supervised truth discovery
approach is proposed. An initial set of known ground truth labels is used to estimate the reliability of sources. The formalization of mutual exclusivity and mutual support between claimed values are exploited to capture the relations between values and to guide the algorithm towards reliability and truth estimations.
Active learning approaches~\cite{yan2011active,bachrach2012grade} address the general
problem of optimally choosing the next instances to label by experts; experts should be only asked if really needed. For example, the authors of~\cite{yan2011active} propose a probabilistic multi-labeler model that enables learning from multiple annotators with varying expertise levels across the information space and provide an optimization formulation for selecting the most uncertain sample and the most reliable annotator to ask for that sample. In~\cite{bachrach2012grade}, the authors propose a probabilistic graphical
model that jointly estimates the difficulties of test questions, the abilities of test participants and the correct answers to questions in aptitude testing
and crowdsourcing settings. The authors devise a scheme for the greedy minimization of expected
model entropy, which allows dynamically choosing
the next question to be asked based on the previous responses. 

Note that LTD approaches do not focus on optimizing the allocation of resources or incorporating a feedback loop, which is typical for active learning scenarios. Hence, although semi-supervised LTD and active learning do have some commonalities, they differ with respect to important aspects. Our approach is designed to deal with the typical LTD setting. 

Despite the variety of LTD techniques and their applications found in research literature, as we will see in the following sections, issues concerning the general practical viability (like efficiency and robustness) are still open. Our approach, LTD-RBM, addresses these issues in a holistic manner and achieves an overall performance that outperforms state-of-the-art techniques on various heterogeneous datasets.


\section{Truth Discovery Formulation}
\label{sec:model}
The aim of latent truth discovery is to identify the underlying truth of facts based on potentially conflicting claims from multiple sources. While all LTD algorithms agree on this general goal, they sometimes differ in the specific definition of facts, claims and sources. Therefore, in this section, we start by formally describing a data model for the LTD task. Afterwards, we will formulate the LTD problem based on the introduced data model.

\begin{table}%
\small 
\begin{tabularx}{\columnwidth}{c|p{10mm}lX}
	\hline\hline
	&Term & Formula & Description\\
	\hline
	&Fact & $f\in\facts$ & A fact can be true or false.\\
	&Truth & $t_f\in\{0,1\}$ & The latent truth whether a fact $f$ is true or false.\\
	\parbox[t]{2mm}{\multirow{4}{*}[2.5em]{\begin{sideways}Data Model\end{sideways}}}
	&Source & $s\in\sources$ & A source that makes claims.\\
	&Claim & $(s,f,v_s^f)\in\claims$ & A positive ($v_s^f = 1$) or negative ($v_s^f = 0$) claim of source $s$ about a fact $f$.\\
	\hline
	&Plausibility & $p_f\in\mathbb{R}$ & Plausibility of a fact $f$ to be true.\\
	&Reliability & $r_s$ & Reliability of a source $s$.\\
	& & $\tpr_s,\fpr_s$ & True / False positive rate.\\
	\parbox[t]{2mm}{\multirow{4}{*}{\begin{sideways}LTD\end{sideways}}}
	&Prevalence & $p_T$ & Probability of a fact to be true.\\
	&Claimed truth & $v_s^f\in\{0,1\}$ & The claimed truth of source $s$ about fact $f$.\\
	&Claiming Sources & $ S_f \subseteq \sources$ & Set of all sources that provide a claim about fact $f$.\\
	&Hidden Truth & $h\in[0,1]$ & $P(h|\vec{v}^f)$ is the estimation of $p_f$ by the LTD-RBM algorithm.\\
	\hline \hline
\end{tabularx}
\vspace{1mm}
\caption{Terminology}
\label{tab:terminology}
\vspace{-5mm}
\end{table}

\subsection{Data Model}
\label{sec:data_model}

For this work, we focus on a \emph{binary model}, \ie a model with facts that can be true or false. 
Thereby, we closely follow models used in multiple previous works~\cite{zhao2012bayesian, galland2010corroborating, wang2012truth}. Each source can make claims about some or all facts, where each claim either states that a fact is true or false.

Formally, we have a set $\facts$ of facts and a set $\sources$ of sources.
Each fact $f\in\facts$ is a combination of a statement together with a Boolean latent truth $t_f\in\{0,1\}$ that represents whether the statement is true or false. For the sake of simplicity, in the following we will refer to a fact $f$ as true or false, depending on $t_f$.
A claim of a source $s\in\sources$ for a fact $f\in\facts$ is then a triple $(s,f,v_s^f)\in\claims\subseteq\sources\times\facts\times\{0,1\}$, where $\claims$ is the set of all claims and $v_s^f\in\{0,1\}$ represents the view of source $s$ on $f$, i.e. whether $s$ considers $f$ to be true or false. 
In this work, we call a claim that a fact is true, a positive claim. In the same manner, claims that a fact is false are called negative claims.
A list of the terminology used in this paper is shown in Tab.~\ref{tab:terminology}.
Note that we do not require sources to provide claims for all facts. 
Table~\ref{tab:Example Claims} shows some example claims to illustrate the model. Each row corresponds to a claim, while facts and sources can be found in the first and second column.


\paragraph{Categorical model}
In many latent truth discovery applications, sources do not make claims about the truth of facts, but claim values for the attributes of entities.
As stated in previous works, such data models are interchangeable~\cite{galland2010corroborating,zhao2012bayesian,pasternack2013latent}: 
Each pair of entity attribute and value can be seen as a fact, and a source claiming a value X can be understood as a positive claim for the fact \emph{entity attribute has value X} and negative claims all other facts for the same attribute of the same entity.
In our example of Tab.~\ref{tab:Example Claims}, an entity would be \emph{Germany}, the attribute would be \emph{Capital of} and city names would be the values. Thus, a claimed value of ``Berlin'', for the \emph{Capital of Germany}, would result in a positive claim for the fact ``Berlin is the Capital of Germany'', but also in negative claims for facts like ``Frankfurt is the Capital of Germany''. 

\begin{table}[bt]
\centering
\small
\begin{tabular}{lll}
\hline\hline
Fact ($\in\facts$)&Source ($\in\sources$)&Claimed Value\\
\hline
Berlin is the Capital of Germany&A&true\\
Berlin is the Capital of Germany&B&true\\
Sydney is the Capital of Australia&A&false\\
Sydney is the Capital of Australia&C&true\\
\dots&\dots&\dots\\
\hline\hline
\end{tabular}
\vspace{1mm}
\caption[Example claims]{Example claims from a geographic domain.}
\label{tab:Example Claims}
\vspace{-5mm}
\end{table}

Note that it depends on the domain whether each entity attribute can only have one true value (like in our example above) or multiple true values (like the author of an article). For further details on how to translate claimed values into logical claims according to the model, we refer to the aforementioned works.

Using our fact-based model (irrespective of whether the original data is fact-based or value-based) we can now formulate the truth discovery problem.

\subsection{Truth Discovery}
\label{sec:def:truth-discovery}
In the following, we will formally define latent truth discovery in terms of our data model. This includes the available input data as well as the two typical goals of LTD: an estimation of the truth of facts and the reliability of sources.

\paragraph{Input}
Initially, all available data is the information given from the sources. That means the input for LTD algorithms is the set $\claims$ of claims, which implicitly contains the sets of facts $\facts$ and sources $\sources$. Based on this information, an LTD algorithm has to perform the inference, typically in an unsupervised fashion.

\paragraph{Truth Estimation} The main goal of latent truth discovery is to distinguish between true and false facts, given the set of claims. Here, we aim for a probabilistic result, \ie we try to find for each fact $f$ a probability $p_f\in[0,1]$ for the fact to be true. 
This probability can be written as 
\begin{equation}
\label{eqn:def:confidence}
	p_f=P(t_f=1|\claims)
\end{equation} 
Note that $t_f$ can depend on all claims instead of just depending on the claims about the fact $f$. The reason is, that $t_f$ also depends on the reliability of sources, which in turn can depend on all claims.

\paragraph{Reliability Estimation} One key for discovering the underlying truth is to distinguish between good and bad sources. To do so, the reliability of sources has to be measured. The range, dimension and interpretation of reliabilities are algorithm-dependent. For example, they can be custom scores~\cite{yin2008truth}, error measures~\cite{galland2010corroborating} or a combination of true positive and false positive rates~\cite{zhao2012bayesian,wang2012truth}. In general, the reliability of a source $s\in\sources$ will be denoted as $r_s$. For our algorithm, $r_s$ is a combination of true positive and false positive rates. Thus, we write $r_s=(\tpr_s,\fpr_s)$.

Note that if we knew the true reliability of each source -- in terms of its true positive rate and true negative rate -- we could aggregate the source claims in a probabilistically optimal way as stated at \cite{wang2012truth,kasneci2017eyetracking}. The only requirement is, that the sources make independent claims.
So the goal for any LTD technique should be to estimate the $\tpr_{s}$ and $\fpr_{s}$ of sources as accurately as possible from given claims. 

\section{Restricted Boltzmann Machines for Latent Truth Discovery}
\label{sec:algorithm}
In this section, we will introduce our specific LTD algorithm, that operates on the previously introduced data model.
This algorithm uses Restricted Boltzmann Machines (RBMs) to learn both, the plausibility of the hidden truth and the reliability of sources in terms of true positive rate (tpr) and false positive rate (fpr). 
We will first give a brief overview of RBMs and their capabilities and will then introduce our specific LTD-RBM model for truth discovery. 

In a first step, this will be done for the case that all sources make claims for all facts. 
Afterwards, we will show how prior knowledge about the sources can be used to initialize our RBMs.
Formulas developed in this section are then used to generalize our approach for cases where sources are not required to make claims for every fact. In the end of this section, we will extend our approach for the usage with categorical data.

\subsection{Restricted Boltzmann Machines}
\label{sec:rbm}
Restricted Boltzmann Machines are a type of neural networks that can be used to learn distributions and hidden factors of observable variables.
In their original form, all units of RBMs are binary variables, although other units are possible~\cite{hinton2012rbms}. In this work, we will concentrate on RBMs with binary variables. 

An RBM consists of two layers that form an undirected, bipartite graph. These two layers can represent observable (visible) variables and hidden factors respectively.
Each edge has a weight that models the mutual influence of the connected visible and hidden unit. There is also no direct influence between the units of the same layer.

This allows for the following probabilistic view on RBMs that considers each unit as a (Bernoulli-distributed) random variable. Given the units of one layer, the units of the other layer are then conditionally independent of each other. Furthermore, the conditional distributions of one layer can be computed from the weights of the RBM and the fixed states of the other layer. Thus, a trained RBM (that has learned the distributions of the training data) allows the computation of the distribution of the hidden factors, given a test example for the visible layer.

The common form for computing these distributions is to use the logistic function $\sigma(x) = \frac{1}{1+e^{-x}}$. With the visible units $\vec{v}=(v_i)_{i=1,...m}$ and the hidden units $\vec{h}=(h_j)_{j=1,...n}$, the conditional probabilities are given by:
\begin{align}
	P(v_i=1|\vec{h}) &= \sigma\left(a_i + \sum\nolimits_{j=1}^nw_{ij}h_j\right)  \label{eqn:prob_visible}\\
	P(h_j=1|\vec{v}) &= \sigma\left(b_j + \sum\nolimits_{i=1}^mw_{ij}v_i\right)  \label{eqn:prob_hidden}
\end{align}
with the weights $w_{ij}$, the visible biases $a_i$ and the hidden biases $b_j$.

In order to be able to compute these probabilities, the RBM has to learn the underlying distributions of a training set of examples for the visible units. This can be done in an unsupervised fashion by feeding the RBM with all observed examples. In our work, we used the Contrastive Divergence (CD) learning method~\cite{hinton2002training} for doing so. CD allows to update the weights for a minibatch of a few facts instead of the whole dataset. Such updating methods are known to show better convergence than global updates and, thus, make the approach more robust.
We refer to previous works~\cite{hinton2012rbms,Smolensky1986rbm,fischer2012introductionRBM} for further details on RBMs.

\subsection{LTD-RBM: RBM-Based Latent Truth Discovery}
\label{sec:our_rbms}
In our work, we utilize RBMs to model the LTD process and for inference (i.e., the discovery of fact truths and source reliabilities). The motivation for this idea is the ability of RBMs to learn hidden factors. Given the aim of sources to provide correct claims, we would expect the main hidden factor behind the claims of all sources is the unknown truth, which we try to discover. 

Following this idea, we design an RBM with one visible unit for each source and only one hidden unit for the hidden truth.
This will allow us to feed the visible layer with information from the claims about one fact and use the hidden unit for reasoning about the hidden truth.
A positive claim from a source can then be encoded by $1$ in the visible unit and a negative claim by a $0$. 

Note that RBMs require the units of one layer to be independent, given the units of the other layer.
For our model, this means that the sources make independent claims, given the hidden truth. This is a common assumption~\cite{zhao2012bayesian,wang2012truth,galland2010corroborating}, although real-world data often contains dependencies between sources, such as copying from each other. In section~\ref{sec:experiments}, we will show that our method still performs well when there are dependencies between the sources, but in the following, we will at first assume an independence of the sources.

Let us for the moment assume that all sources make claims about a fact $f$. 
We then get an input vector $\vec{v}^f$ for the visible layer from all claims made about this fact. 
For the sake of simplicity, we will assume $\sources=\{s_1,\ldots,s_n\}$ and identify the claim of $s\in\sources$ about the fact $f$ with $v_s^f\in\{0,1\}$.

Thus, in order to learn the underlying distribution of our data, we can train our RBM iteratively with the input vectors $v_s^f$ of all facts, using Contrastive Divergence as mentioned above. We perform this repeatedly with all facts until the weights in the RBM converge.

In a trained model, which has learned the underlying distribution, the probability $p_f$ that a fact $f$ is true, given the set of all claims about the fact, is then equivalent to the probability that the hidden unit is 1, given the corresponding visible units. 
This probability is defined in equation (\ref{eqn:prob_hidden}) and can directly be computed. Note that our model has only one hidden unit, so we can omit $j$. In this way, we get:
\begin{equation}
	p_f=P(h|\vec{v}^f)
	\label{eqn:plausibility}
\end{equation}

Furthermore, the reliability of sources can also be expressed in terms of the Restricted Boltzmann Machine. 
The true positive rate of a source $s$ is the probability that $s$ makes a positive claim for a fact under the condition of the fact to be true.
A positive fact corresponds to the hidden layer being $1$ and a positive claim of $s$ corresponds to the visible unit $v_s$ to be $1$. Analogously, we can express negative facts and claims. This leads to
\begin{align}
	\tpr_s &= P(v_s=1|h=1) = \sigma\left(a_s + w_s\right)
	\label{eqn:rbm_tpr}\\
	\fpr_s &= P(v_s=1|h=0) = \sigma\left(a_s\right)
	\label{eqn:rbm_fpr}
\end{align}
Using these equations, we can see that a trained RBM solves the truth discovery problem. That means that truth discovery requires training an RBM with the given claims. Afterwards, the plausibility and the source reliability can directly be computed as shown.

\subsection{Prior Knowledge}
In the previous subsection, we dealt with trained RBMs and showed how they can be used for truth discovery. 
For doing so, the RBMs have to be trained with the given set of claims. Basically, the training data can also be explained by a dual model:
Replacing the hidden truth $H$ with its inverse $H^* := 1 - H$ would swap true and false positive rate of all sources, but explain the model in the same way. Thus, the training can lead to different models. 

Under the assumption that most sources aim at providing correct information, a smart initialization can ensure that the training process creates the desired model. Furthermore, initializing the RBM close to the real distribution will reduce the required training steps. For this purpose, we will show how an initial estimation of the true and false positive rates can be used for initialization. Often, these initial estimations are just based on the confidence that most sources are good, which leads to the same initial values $\tpr > 0.5$ and $\fpr < 0.5$ for all sources. Subsequent training will adjust these initial beliefs to the real data. Our method also allows for more specific initializations, e.g. different values for different types of sources. It is even possible to initialize the model with well-known individual values for the sources and skip the training that has the goal of finding these values.
In order to find the initial weights for the RBM, we start with equations (\ref{eqn:rbm_tpr}) and (\ref{eqn:rbm_fpr}):
\begin{align}
	\nonumber \fpr_s &\stackrel{(\ref{eqn:rbm_fpr})}= \sigma(a_s) = \frac{1}{1 + e^{-a_s}}\\
	\label{eqn:resolve_ai}\Rightarrow a_s & = - \log\left(\frac{1}{\fpr_s}-1\right) = \log\left(\frac{\fpr_s}{1-\fpr_s}\right)\\
	\nonumber \tpr_s &\stackrel{(\ref{eqn:rbm_tpr})}= \sigma(a_s+w_s) = \frac{1}{1 + e^{-a_s-w_s}}\\
	\label{eqn:resolve_wi}\Rightarrow w_s &= \log\left(\frac{\tpr_s}{1-\tpr_s}\right) - \log\left(\frac{\fpr_s}{1-\fpr_s}\right)
\end{align}
It is also possible to compute the bias $b$ for the hidden unit. Using the prevalence $p_T = P(h=1)$, we get:
\begin{equation}
  \label{eqn:resolve_b}
	b = \log\left(\frac{p_T}{1-p_T}\right) + \sum_s\left(\log(1-\tpr_s) - \log(1-\fpr_s)\right)
\end{equation}
The derivation for this formula can be found in appendix \ref{app:bias}. 

Based on these formulas, we can initialize the setting of source reliabilities in an LTD-RBM, if the prevalence is given. 
This will be used in the next subsection to deal with facts that do not have claims from all sources.

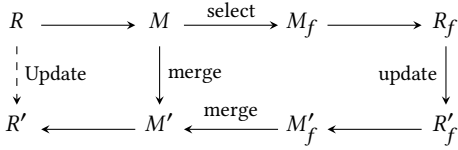
\begin{figure}%
\centering
\usetikzlibrary{matrix}
\begin{tikzpicture}
\matrix (m) [matrix of math nodes,row sep=2.5em,column sep=4em,minimum width=2em]
{
	 R_{\vphantom{f}} & M_{\vphantom{f}} &  M_f & R_f\\
	 R'_{\vphantom{f}} & M'_{\vphantom{f}} &  M'_f & R_f'\\};
\path[-stealth]
	(m-1-1) edge [dashed] node [right] {\small Update} (m-2-1)
	        edge (m-1-2)
  (m-1-2) edge node[above] {\small select} (m-1-3)
	        edge node[right] {\small merge} (m-2-2)
	(m-1-3) edge (m-1-4)
	(m-1-4) edge node[left] {\small update} (m-2-4)
	(m-2-4) edge (m-2-3)
	(m-2-3) edge node[above] {\small merge} (m-2-2)
	(m-2-2) edge (m-2-1);
\end{tikzpicture}
\vspace{-7mm}
\caption[Update in case of missing claims]{Update in case of missing claims: The weights of an RBM $R$ are updated into weights $R'$ by switching to the model $M$ and using a helper $R_f$ for the actual update.}
\label{fig:update_missing}%
\end{figure}

\subsection{Missing data}
Using the previous insights, we can see from (\ref{eqn:rbm_tpr}) and (\ref{eqn:rbm_fpr}) how to compute the source qualities from the weights of an RBM. In addition (\ref{eqn:resolve_ai}), (\ref{eqn:resolve_wi}) and (\ref{eqn:resolve_b}) show how to compute the weights of an RBM from the source qualities. This leads to a bijection that we will use to deal with facts that do not have claims from all sources.

The idea is to construct for each fact $f$ a helping RBM that operates only on sources with claims for fact $f$. This new RBM allows performing all operations for $f$ as described in subsection \ref{sec:our_rbms}. For the training operation, the updated weights can be transferred back to the original RBM. 
Two operations are affected by missing claims: training an RBM and computing the plausibility. In the following, we will describe in detail how the helping RBM can be used to perform them.

Let $S_f\subseteq\sources$ be the set of sources that provide a claim about a fact $f$. As mentioned before, we want to create a helping RBM that only has visible units for the sources in $S_f$. For that purpose, we identify a given RBM $R$ with the corresponding model of the sources $M=\left(p_T,(\tpr_s)_{s\in\sources},(\fpr_s)_{s\in\sources}\right)$. It is then possible to shrink $M$ to a model $M_f$ by only using sources from $S_f$. $M_f$ can again be identified with an RBM $R_f$, which is the desired helping RBM. This process is illustrated in Fig.~\ref{fig:update_missing}.

It is now possible to compute the probability of $f$ being true by using $R_f$. Keep in mind that this is based on the model $M_f$, which is a sub-model of $M$. Thus, the computed probability is consistent with the learned distribution of $R$.
During the training, $R_f$ can be updated into a new RBM $R_f'$. This RBM corresponds bijectively to an updated model $M_f'$, this can be extended to a model $M'$ by filling the rates of the missing sources with the (unchanged) rates from the original model $M$. See Fig. \ref{fig:update_missing} for the complete chain of updating an RBM $R$ into $R'$ when not all sources provide claims for $f$.


\begin{figure}%
\begin{tikzpicture}[scale=0.28]
\node at (6,0.5) {\bf Claims};
\node at (13,0.5) {\bf Truth};

\node[thick,circle,draw, minimum size=1] (V1) at  (6,-2) {$v_1$};
\node[thick,circle,draw, minimum size=1] (V2) at  (6,-6) {$v_2$};
\node[thick,circle,draw, minimum size=1] (V3) at  (6,-10) {$v_3$};

\node[thick,draw, minimum size=1] (a1) at  (3,-2) {$a_1$};
\node[thick,draw, minimum size=1] (a2) at  (3,-6) {$a_2$};
\node[thick,draw, minimum size=1] (a3) at  (3,-10) {$a_3$};

\node[thick,draw, minimum size=1] (b) at  (13,-9) {$b$};

\node[thick,circle,draw, minimum size=1] (H) at  (13,-6) {$h$};

\draw [thick] (V1) -- (H) node[midway, above]{$w_{1}$};
\draw [thick] (V2) -- (H) node[midway, above]{$w_{2}$};
\draw [thick] (V3) -- (H) node[midway, above]{$w_{3}$};

\draw [thick] (V1) -- (a1);
\draw [thick] (V2) -- (a2);
\draw [thick] (V3) -- (a3);

\draw [thick] (H) -- (b);

\node at (9.5,-12) {\bf (a)};
\node at (24.5,-12) {\bf (b)};

\node at (21,0.5) {\bf Claims};
\node at (28,0.5) {\bf Truth};

\node[thick,circle,draw, minimum size=1] (V1-2) at  (21,-2) {$v_1$};
\node[thick,circle,draw, minimum size=1] (V2-2) at  (21,-6) {$v_2$};
\node[thick,circle,draw, minimum size=1] (V3-2) at  (21,-10) {$v_3$};

\node[thick,draw, minimum size=1] (a1-2) at  (18,-2) {$a_1$};
\node[thick,draw, minimum size=1] (a3-2) at  (18,-10) {$a_3$};

\node[thick,draw, minimum size=1] (b-2) at  (28,-9) {$b_f$};

\node[thick,circle,draw, minimum size=1] (H-2) at  (28,-6) {$h$};

\draw [thick] (V1-2) -- (H-2) node[midway, above]{$w_{1}$};
\draw [thick] (V3-2) -- (H-2) node[midway, above]{$w_{3}$};

\draw [thick] (19.7,-4.7) -- (22.3,-7.3);
\draw [thick] (19.7,-7.3) -- (22.3,-4.7);

\draw [thick] (V1-2) -- (a1-2);
\draw [thick] (V3-2) -- (a3-2);

\draw [thick] (H-2) -- (b-2);

\node at (13,0.5) {\bf Truth};

\end{tikzpicture}
\vspace{-7mm}
\caption[Update in case of missing claims]{Example for missing claims: The weights of an RBM $R$ are transformed into weights of $R_f$ with less input units (i.e. less sources).}
\label{fig:update_missing_example}%
\end{figure}
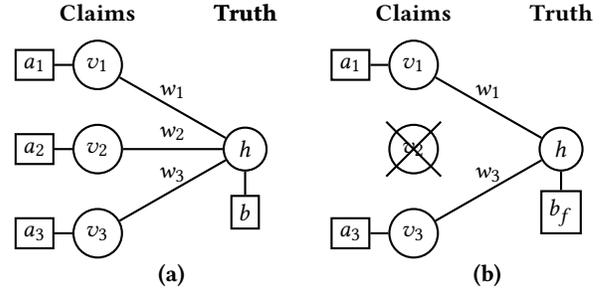

\paragraph{Implementation} Computing a new RBM for each fact is resource consuming and inefficient. In order to avoid this huge overhead, we do not explicitly construct a new RBM, but implemented the RBMs in a way that allows to act like the helping $R_f$.
The bias $a_i$ and the weight $w_i$ only depend on the true and false positive rates of source $s_i$ (see equations (\ref{eqn:resolve_ai}) and (\ref{eqn:resolve_wi})). This means that removing a source does not affect these values and only the bias $b$ must be adjusted in case of missing claims. Figure \ref{fig:update_missing_example} shows an example for the result.

In equation \ref{eqn:resolve_b} $b$ can be split into $b := b_G + \sum_sb_s$ with a global part $b_G$ and a source dependent part $b_s$:
\begin{align*}
	b_G &:= \log\left(\frac{p_T}{1-p_T}\right)\\
	b_s &:= \left(\log(1-\tpr_s) - \log(1-\fpr_s)\right)
\end{align*}
This allows to easily adjust $b$ to missing claims by removing the corresponding source-dependent parts.
When updating the weight $b$ into $b'$, we compute the updated source dependent biases $b_s'$ from the updated true / false positive rates. This allows us to compute the difference for the global bias, which allows us to update the global bias.

Note that it is possible to skip the steps initialization and handling of missing data (by filling in negative claims). However, we observed in experiments, that without initialization we obtained random results and without a treatment for missing data we obtained results of poor quality. Hence, in the following all experiments included these steps.

\subsection{Categorical data}
\label{subsec:categorical-rbm}
In the previous parts, we concentrated binary data where facts can either be true or false. 
Often, datasets contain categorical data, such as dates or names. In the categorical case, sources claim values for attributes of entities.
As mentioned in section \ref{sec:data_model}, multinomial data can be transformed into binary data by creating one binary fact for each claimed value.

In the following, we will consider the case where only one value can be correct per entity attribute. The above binarization of categorical data has the disadvantage that the resulting facts are independent and the constraint of having only one correct value is lost. In the following, we will show how this constraint can be included again. First this is done as a post-processing step for any probabilistic truth discovery algorithm, which allows to adjust the computed plausibilities. Afterwards, we will incorporate the constraint into our algorithm, which results in both modified learning and prediction methods.

We start by formalizing the categorical model:
\subsubsection*{Definitions}
Given a set $\mathcal{A}$ of entity attributes, for each attribute $a\in\mathcal{A}$ we denote the set of sources that make a claim about $a$ as $S_a\in\sources$ and the set of claimed values as $\Omega_a$. When binarizing the data, we get one fact $f = (a,c)$ for each attribute $a\in\mathcal{A}$ and each value $c\in \Omega_a$.

This leads to the binary model as described in \ref{sec:data_model} in which we now denote a fact by the pair $(a,c)$ (and thus replace $f$ in the terms). A claim in the binary model is positive, i.e. $v_s^{(a,c)} = 1$, if a the source $s$ claims that $c$ is the value of $a$.

\begin{figure}[bt]%
\begin{tikzpicture}
\begin{axis}[
    ylabel = {\small Time (s)},
		xlabel = {\small Number of Claims},
		legend pos = north west,
		legend style={font=\small},
		xtick={2,3,4,5,6},
    xticklabels={$10^2$,$10^3$,$10^4$,$10^5$,$10^6$},
		y post scale=.6,
]
\addplot[
    color = {red},
]
coordinates {
    (2.199121,0.000210) +- (0.000000,0.000058)
    (2.369477,0.000283) +- (0.000000,0.000075)
    (2.539832,0.000352) +- (0.000000,0.000070)
    (2.710188,0.000573) +- (0.000000,0.001707)
    (2.880544,0.000722) +- (0.000000,0.001524)
    (3.050899,0.001030) +- (0.000000,0.002148)
    (3.221255,0.001370) +- (0.000000,0.002664)
    (3.391611,0.001959) +- (0.000000,0.001956)
    (3.561966,0.003140) +- (0.000000,0.004400)
    (3.732322,0.004606) +- (0.000000,0.003610)
    (3.902678,0.007406) +- (0.000000,0.006629)
    (4.073033,0.010871) +- (0.000000,0.004589)
    (4.243389,0.017191) +- (0.000000,0.004222)
    (4.413744,0.026834) +- (0.000000,0.006420)
    (4.584100,0.041880) +- (0.000000,0.009892)
    (4.754456,0.062147) +- (0.000000,0.012840)
    (4.924811,0.091987) +- (0.000000,0.017978)
    (5.095167,0.137602) +- (0.000000,0.025754)
    (5.265523,0.198565) +- (0.000000,0.037717)
    (5.435878,0.289536) +- (0.000000,0.040914)
};
\addlegendentry{MajorityVoting}
\addplot[
    color = {blue},
]
coordinates {
    (2.199121,0.000543) +- (0.000000,0.000121)
    (2.369477,0.000725) +- (0.000000,0.000140)
    (2.539832,0.000955) +- (0.000000,0.000221)
    (2.710188,0.001426) +- (0.000000,0.001773)
    (2.880544,0.001858) +- (0.000000,0.000545)
    (3.050899,0.002774) +- (0.000000,0.001058)
    (3.221255,0.004109) +- (0.000000,0.001427)
    (3.391611,0.006723) +- (0.000000,0.004388)
    (3.561966,0.010870) +- (0.000000,0.006653)
    (3.732322,0.018104) +- (0.000000,0.010590)
    (3.902678,0.030295) +- (0.000000,0.016348)
    (4.073033,0.052491) +- (0.000000,0.028503)
    (4.243389,0.088954) +- (0.000000,0.045225)
    (4.413744,0.141048) +- (0.000000,0.062096)
    (4.584100,0.227069) +- (0.000000,0.109831)
    (4.754456,0.329220) +- (0.000000,0.107900)
    (4.924811,0.498904) +- (0.000000,0.171951)
    (5.095167,0.753925) +- (0.000000,0.260703)
    (5.265523,1.188458) +- (0.000000,0.429392)
    (5.435878,1.599797) +- (0.000000,0.383113)
};
\addlegendentry{LTD-RBM}
\addplot[
    color = {green},
]
coordinates {
    (2.199121,0.000435) +- (0.000000,0.000157)
    (2.369477,0.000677) +- (0.000000,0.000335)
    (2.539832,0.001013) +- (0.000000,0.000523)
    (2.710188,0.001470) +- (0.000000,0.000730)
    (2.880544,0.002173) +- (0.000000,0.001075)
    (3.050899,0.003371) +- (0.000000,0.001689)
    (3.221255,0.005188) +- (0.000000,0.002780)
    (3.391611,0.008213) +- (0.000000,0.004011)
    (3.561966,0.012803) +- (0.000000,0.005870)
    (3.732322,0.020853) +- (0.000000,0.010009)
    (3.902678,0.033588) +- (0.000000,0.017060)
    (4.073033,0.058536) +- (0.000000,0.031368)
    (4.243389,0.096118) +- (0.000000,0.047896)
    (4.413744,0.154899) +- (0.000000,0.074862)
    (4.584100,0.253768) +- (0.000000,0.120163)
    (4.754456,0.392425) +- (0.000000,0.179964)
    (4.924811,0.583089) +- (0.000000,0.247759)
    (5.095167,0.876373) +- (0.000000,0.345899)
    (5.265523,1.178270) +- (0.000000,0.429020)
    (5.435878,1.715576) +- (0.000000,0.748091)
};
\addlegendentry{MLE}
\addplot[
    color = {purple},
]
coordinates {
    (2.199121,0.003266) +- (0.000000,0.000421)
    (2.369477,0.004812) +- (0.000000,0.002563)
    (2.539832,0.006152) +- (0.000000,0.001866)
    (2.710188,0.008544) +- (0.000000,0.002331)
    (2.880544,0.011987) +- (0.000000,0.003185)
    (3.050899,0.018685) +- (0.000000,0.005247)
    (3.221255,0.027979) +- (0.000000,0.008516)
    (3.391611,0.044022) +- (0.000000,0.012104)
    (3.561966,0.068283) +- (0.000000,0.017583)
    (3.732322,0.108768) +- (0.000000,0.030392)
    (3.902678,0.176740) +- (0.000000,0.049730)
    (4.073033,0.279892) +- (0.000000,0.075010)
    (4.243389,0.448218) +- (0.000000,0.108764)
    (4.413744,0.708311) +- (0.000000,0.188481)
    (4.584100,1.097604) +- (0.000000,0.250877)
    (4.754456,1.630812) +- (0.000000,0.362610)
    (4.924811,2.420866) +- (0.000000,0.525697)
    (5.095167,3.708150) +- (0.000000,0.795255)
    (5.265523,5.490817) +- (0.000000,1.235514)
    (5.435878,8.034468) +- (0.000000,1.434619)
};
\addlegendentry{LTM}
\addplot[
    color = {orange},
]
coordinates {
    (2.199121,0.000627) +- (0.000000,0.000259)
    (2.369477,0.001036) +- (0.000000,0.000372)
    (2.539832,0.001541) +- (0.000000,0.000710)
    (2.710188,0.002365) +- (0.000000,0.001130)
    (2.880544,0.003622) +- (0.000000,0.001742)
    (3.050899,0.005838) +- (0.000000,0.003166)
    (3.221255,0.009188) +- (0.000000,0.005836)
    (3.391611,0.015144) +- (0.000000,0.008405)
    (3.561966,0.023627) +- (0.000000,0.012651)
    (3.732322,0.041248) +- (0.000000,0.023984)
    (3.902678,0.068230) +- (0.000000,0.040143)
    (4.073033,0.111287) +- (0.000000,0.065166)
    (4.243389,0.197361) +- (0.000000,0.121014)
    (4.413744,0.301276) +- (0.000000,0.174731)
    (4.584100,0.477118) +- (0.000000,0.270670)
    (4.754456,0.751710) +- (0.000000,0.491684)
    (4.924811,1.138118) +- (0.000000,0.676015)
    (5.095167,1.707588) +- (0.000000,1.142199)
    (5.265523,2.643863) +- (0.000000,1.797644)
    (5.435878,3.695996) +- (0.000000,2.148070)
};
\addlegendentry{2-Estimates}
\end{axis}
\end{tikzpicture}%
\caption{Computational time in relation to the number of claims in a dataset.}%
\label{fig:time_vs_claims}%
\vspace{-2mm}
\end{figure}
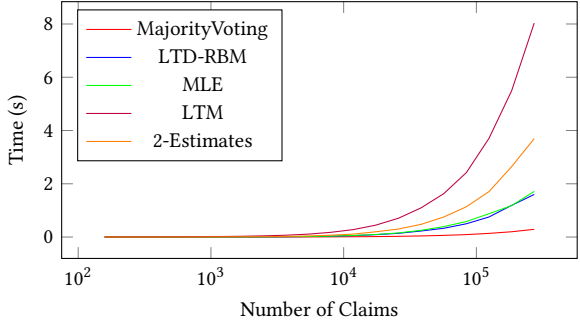

\subsubsection*{Post-Processing Step}
Given a probabilistic truth discovery algorithm, we get for each fact $f$ a probability that $f$ is true. Following the notion of (\ref{eqn:def:confidence}), we get for each fact $f=(a,c)$ a value
\begin{equation}
\label{eqn:def:confidence_multi}
p_{(a,c)} = P\left(t_{(a,c)} = 1 \middle| \claims\right)
\end{equation}
As mentioned above, this formula does not include the knowledge that a categorical attribute takes exactly one value. This knowledge can be expressed by the constraint $\sum_{c\in\Omega_a}t_{(a,c)} = 1$. Including this into (\ref{eqn:def:confidence_multi}) leads to the following conditioned probability:
\begin{align}
	p_{(a,c)}^* & = P\left(t_{(a,c)} = 1 \middle| \sum\nolimits_{c'\in\Omega_a}t_{(a,c')} = 1;\claims\right)\\
			& = \frac{P\left(t_{(a,c)} = 1 ,\sum\nolimits_{c'\in\Omega_a}t_{(a,c')} = 1\middle|\claims\right)}{P\left(\sum\nolimits_{c'\in\Omega_a}t_{(a,c')} = 1\middle|\claims\right)}   \nonumber\\
			& = \frac{p_{(a,c)}\prod_{c'\in\Omega_a\atop c'\not=c}(1-p_{(a,c')})}{P\left(\sum\nolimits_{c'\in\Omega_a}t_{(a,c')} = 1\middle|\claims\right)}   \nonumber\\
			& = \frac{\frac{p_{(a,c)}}{1-p_{(a,c)}}\prod_{c'\in\Omega_a}(1-p_{(a,c')})}{\sum_{d\in\Omega_a}\frac{p_{(a,d)}}{1-p_{(a,d)}}\prod_{c'\in\Omega_a}(1-p_{(a,c')})} \nonumber\\
			& = \frac{\frac{p_{(a,c)}}{1-p_{(a,c)}}}{\sum_{d\in\Omega_a}\frac{p_{(a,d)}}{1-p_{(a,d)}}}
\end{align}

This mean that normalizing the odds over the different values for an entity attribute leads to the desired adjusted plausibility values. Note that this adjustment does not change the order of the confidences: the value $c$ with the highest confidence $p_{(a,c)}$ has also the highest confidence $p_{(a,c)}^*$.

\subsubsection*{Modified LTD-RBM}
While the post-processing step works with every latent truth discovery algorithm that provides probabilistic plausibility values, it does not change the order of the plausibilites. Thus, if always the value with the highest plausibility is considered as the true value, the adjustment of the plausibilities does not change the outcome.

In the following, we show how above results can be used to modify our RBM-based algorithm. This allows to include the constraint of exactly one correct value per entity attribute into the training of the RBM. We begin by computing the odds of the plausibility for the RBM model:
\begin{align*}
	\frac{p_{(a,c)}}{1-p_{(a,c)}} &= \frac{\sigma\left(T_{(a,c)}\right)}{1-\sigma\left(T_{(a,c)}\right)}
		= \frac{\frac{1}{1 + e^{-T_{(a,c)}}}}{\frac{e^{-T_{(a,c)}}}{1 + e^{-T_{(a,c)}}}}
		= e^{T_{(a,c)}}
\end{align*}
with 
\[T_{(a,c)} := b + \sum_{s\in S_a} v_s^{(a,c)}w_s\]
Using these results, we can express the adjusted plausibilities by a softmax function:
\[
p_{(a,c)}^* = \frac{e^{T_{(a,c)}}}{\sum_{d\in\Omega_a}e^{T_{(a,d)}}}
\]
Using this formula, it is possible to modify the contrastive divergence for learning the weights of the RBM. Contrastive divergence requires to Bernoulli-sample the hidden state based on the current probability $p_{(a,c)}$. By using the adjusted probability instead, we get a categorical inference method for the truth discovery problem.

\section{Evaluation}
\label{sec:experiments}
In this section, we will evaluate how well LTD-RBM performs and compare it with state-of-the-art truth discovery algorithms. 
We examine different aspects of the algorithms, including their effectiveness, efficiency and robustness.
For this purpose, we utilize synthetic datasets in \ref{sec:eval_synthetic} as well as real-world datasets in \ref{sec:eval_real}. 
Before, we describe the hyperparameters of the different methods in \ref{sec:hyper_parameters}. These will play a role in \ref{sec:eval_real}, where we optimize the hyperparameter to find good values for the different datasets.

We will compare our method to the \emph{Maximum Likelihood Estimation} (MLE) approach~\cite{wang2012truth}\footnote{The original MLE publication takes all missing claims as negative claims. For our data sets, we adapted the formulas to the case of positive, negative and missing claims.}, the \emph{Latent Truth Model} (LTM) approach~\cite{zhao2012bayesian}, the 2-Estimates method~\cite{galland2010corroborating} and majority voting.

\paragraph{Categorical Data} 
In our experiments, we have to deal with categorical data.
Since all evaluated algorithms expect binary data, we have to transform the data into the binary model, as described in in \ref{subsec:categorical-rbm}. 
Each of these facts gets a plausibility score (of being true) from the LTD-Algorithms. 
For the sake of evaluation, we then take for each entity attribute the fact with the highest probability as the true fact and all other facts are considered to be false.

\subsection{Hyperparameters}
\label{sec:hyper_parameters}
The performance of our algorithm for truth discovery can be influenced by a number of hyperparameters, such as the learning rate or the initial
source reliabilities. In the following, we discuss the hyperparameters of the above mentioned methods. 

\paragraph{Prior Source Reliabilities}
One type of hyperparameters are the initial or prior source reliabilities in terms of initial true and false positive rates and an initial prevalence.
In many applications, no concrete prior source reliability is known.
In that case, choosing a concrete initial/prior source reliability can be difficult and, thus, a low sensitivity towards the choice of these parameters is desirable.

\paragraph{Number of iterations}
Except Majority Voting, all mentioned algorithms perform an iterative truth discovery. The number of iterations influences both the accuracy of the output as well as computational time. Thus, the choice of the relevant parameters has to be done according to a desired trade-off between time and accuracy.
Depending on the algorithm, the number of iteration is influenced by different parameters: LTM uses a Gibbs sampling. Here we use the recommendes number of $50$ iterations. The other algorithms use a convergence criterion to stop the iterations. 

\paragraph{Further Parameters}
Apart from the above mentioned hyperparameters, the training of the Restricted Boltzmann Machine requires a learning rate and a decay factor for it.
Furthermore, 2-Estimates relies on a linearly decreasing weight $\lambda$. We use the number of steps needed for $\lambda$ to decrease from $1$ to $0$ as another hyperparameter.

\paragraph{Choice of Hyperparameters} For the synthetic dataset, we set the parameters to comparable values. That is in our case a tpr of $0.8$ and a fpr of $0.2$. The learning rate for LTD-RBM was set to $0.01\cdot\exp(-0.5 \cdot n)$ in the $n$-th iteration. The weights for the LTM-prior are set according to the original publication ($100$ for $\tpr$, $1010$ for $\fpr$ and $10$ for $p_T$). The 2-Estimates weight $\lambda$ decreases within 10 steps to 0.
For the real-world dataset, we perform a dataset specific optimization, which is explained in~\ref{sec:eval_real}.

\begin{figure*}[bt]%
\begin{tikzpicture}
\begin{groupplot}[
    group style = {group size=1 by 3,vertical sep=0pt},
		x tick label style={font=\tiny},
		y tick label style={font=\tiny},
		x post scale=.75
]
\nextgroupplot[
    ylabel = {\small Accuracy},
    legend pos = south west,
		legend style={font=\tiny},
		y post scale=.65
]
\addplot[
    color = {red},
]
coordinates {
    (0.022977,0.804294) +- (0.000000,0.174804)
    (0.068932,0.764771) +- (0.000000,0.164739)
    (0.114886,0.725977) +- (0.000000,0.181635)
    (0.160841,0.736169) +- (0.000000,0.165304)
    (0.206796,0.722092) +- (0.000000,0.184774)
    (0.252750,0.761306) +- (0.000000,0.175357)
    (0.298705,0.745753) +- (0.000000,0.191107)
    (0.344659,0.739253) +- (0.000000,0.188755)
    (0.390614,0.751767) +- (0.000000,0.187928)
    (0.436568,0.736912) +- (0.000000,0.191574)
    (0.482523,0.737680) +- (0.000000,0.190918)
    (0.528478,0.717319) +- (0.000000,0.181597)
    (0.574432,0.720201) +- (0.000000,0.185948)
    (0.620387,0.701863) +- (0.000000,0.187523)
    (0.666341,0.673292) +- (0.000000,0.179469)
    (0.712296,0.683336) +- (0.000000,0.175763)
    (0.758250,0.658760) +- (0.000000,0.168944)
    (0.804205,0.623059) +- (0.000000,0.164077)
    (0.850160,0.581079) +- (0.000000,0.137839)
    (0.896114,0.545990) +- (0.000000,0.108316)
};
\addlegendentry{MajorityVoting}
\addplot[
    color = {blue},
]
coordinates {
    (0.022977,0.805621) +- (0.000000,0.175468)
    (0.068932,0.767769) +- (0.000000,0.166143)
    (0.114886,0.729648) +- (0.000000,0.184871)
    (0.160841,0.741770) +- (0.000000,0.168448)
    (0.206796,0.727455) +- (0.000000,0.188060)
    (0.252750,0.767347) +- (0.000000,0.178605)
    (0.298705,0.751966) +- (0.000000,0.195145)
    (0.344659,0.745152) +- (0.000000,0.192994)
    (0.390614,0.757594) +- (0.000000,0.192215)
    (0.436568,0.742252) +- (0.000000,0.196652)
    (0.482523,0.743827) +- (0.000000,0.197423)
    (0.528478,0.724795) +- (0.000000,0.189129)
    (0.574432,0.726685) +- (0.000000,0.194115)
    (0.620387,0.708706) +- (0.000000,0.197271)
    (0.666341,0.678640) +- (0.000000,0.192642)
    (0.712296,0.690394) +- (0.000000,0.191714)
    (0.758250,0.668874) +- (0.000000,0.189304)
    (0.804205,0.633991) +- (0.000000,0.191051)
    (0.850160,0.596000) +- (0.000000,0.175649)
    (0.896114,0.551614) +- (0.000000,0.171633)
};
\addlegendentry{LTD-RBM}
\addplot[
    color = {green},
]
coordinates {
    (0.022977,0.805053) +- (0.000000,0.174805)
    (0.068932,0.766772) +- (0.000000,0.165641)
    (0.114886,0.729110) +- (0.000000,0.183808)
    (0.160841,0.740087) +- (0.000000,0.168489)
    (0.206796,0.727143) +- (0.000000,0.188164)
    (0.252750,0.767223) +- (0.000000,0.178320)
    (0.298705,0.751733) +- (0.000000,0.195568)
    (0.344659,0.745341) +- (0.000000,0.193931)
    (0.390614,0.757644) +- (0.000000,0.193440)
    (0.436568,0.742304) +- (0.000000,0.197975)
    (0.482523,0.743147) +- (0.000000,0.200681)
    (0.528478,0.723590) +- (0.000000,0.191961)
    (0.574432,0.726430) +- (0.000000,0.198081)
    (0.620387,0.708570) +- (0.000000,0.202910)
    (0.666341,0.681420) +- (0.000000,0.198287)
    (0.712296,0.695522) +- (0.000000,0.197386)
    (0.758250,0.672962) +- (0.000000,0.201063)
    (0.804205,0.639269) +- (0.000000,0.209615)
    (0.850160,0.605990) +- (0.000000,0.196973)
    (0.896114,0.570082) +- (0.000000,0.217705)
};
\addlegendentry{MLE}
\addplot[
    color = {purple},
]
coordinates {
    (0.022977,0.804959) +- (0.000000,0.175002)
    (0.068932,0.766832) +- (0.000000,0.166065)
    (0.114886,0.729139) +- (0.000000,0.183586)
    (0.160841,0.740648) +- (0.000000,0.168230)
    (0.206796,0.726571) +- (0.000000,0.187524)
    (0.252750,0.766771) +- (0.000000,0.177597)
    (0.298705,0.751345) +- (0.000000,0.194779)
    (0.344659,0.744823) +- (0.000000,0.192107)
    (0.390614,0.758281) +- (0.000000,0.191003)
    (0.436568,0.742085) +- (0.000000,0.196449)
    (0.482523,0.743782) +- (0.000000,0.197965)
    (0.528478,0.723727) +- (0.000000,0.190033)
    (0.574432,0.726667) +- (0.000000,0.194974)
    (0.620387,0.707775) +- (0.000000,0.199753)
    (0.666341,0.678490) +- (0.000000,0.194969)
    (0.712296,0.692424) +- (0.000000,0.193414)
    (0.758250,0.671045) +- (0.000000,0.192769)
    (0.804205,0.636955) +- (0.000000,0.200320)
    (0.850160,0.608607) +- (0.000000,0.186827)
    (0.896114,0.591071) +- (0.000000,0.199920)
};
\addlegendentry{LTM}
\addplot[
    color = {orange},
]
coordinates {
    (0.022977,0.805730) +- (0.000000,0.175256)
    (0.068932,0.767790) +- (0.000000,0.167370)
    (0.114886,0.727703) +- (0.000000,0.184725)
    (0.160841,0.738763) +- (0.000000,0.169363)
    (0.206796,0.720178) +- (0.000000,0.187629)
    (0.252750,0.755553) +- (0.000000,0.174762)
    (0.298705,0.737110) +- (0.000000,0.192862)
    (0.344659,0.728245) +- (0.000000,0.192689)
    (0.390614,0.734241) +- (0.000000,0.190959)
    (0.436568,0.714694) +- (0.000000,0.195733)
    (0.482523,0.711513) +- (0.000000,0.201992)
    (0.528478,0.682931) +- (0.000000,0.197611)
    (0.574432,0.684521) +- (0.000000,0.200801)
    (0.620387,0.652282) +- (0.000000,0.211897)
    (0.666341,0.621177) +- (0.000000,0.210922)
    (0.712296,0.625156) +- (0.000000,0.218004)
    (0.758250,0.613294) +- (0.000000,0.220483)
    (0.804205,0.571806) +- (0.000000,0.229869)
    (0.850160,0.559742) +- (0.000000,0.214692)
    (0.896114,0.554175) +- (0.000000,0.249886)
};
\addlegendentry{2-Estimates}

\nextgroupplot[
		ylabel = {\small std(Accuracy)},
		y post scale=.65
]
\addplot[
    color = {red},
]
coordinates {
    (0.022977,0.174804)
    (0.068932,0.164739)
    (0.114886,0.181635)
    (0.160841,0.165304)
    (0.206796,0.184774)
    (0.252750,0.175357)
    (0.298705,0.191107)
    (0.344659,0.188755)
    (0.390614,0.187928)
    (0.436568,0.191574)
    (0.482523,0.190918)
    (0.528478,0.181597)
    (0.574432,0.185948)
    (0.620387,0.187523)
    (0.666341,0.179469)
    (0.712296,0.175763)
    (0.758250,0.168944)
    (0.804205,0.164077)
    (0.850160,0.137839)
    (0.896114,0.108316)
};
\addplot[
    color = {blue},
]
coordinates {
    (0.022977,0.175468)
    (0.068932,0.166143)
    (0.114886,0.184871)
    (0.160841,0.168448)
    (0.206796,0.188060)
    (0.252750,0.178605)
    (0.298705,0.195145)
    (0.344659,0.192994)
    (0.390614,0.192215)
    (0.436568,0.196652)
    (0.482523,0.197423)
    (0.528478,0.189129)
    (0.574432,0.194115)
    (0.620387,0.197271)
    (0.666341,0.192642)
    (0.712296,0.191714)
    (0.758250,0.189304)
    (0.804205,0.191051)
    (0.850160,0.175649)
    (0.896114,0.171633)
};
\addplot[
    color = {green},
]
coordinates {
    (0.022977,0.174805)
    (0.068932,0.165641)
    (0.114886,0.183808)
    (0.160841,0.168489)
    (0.206796,0.188164)
    (0.252750,0.178320)
    (0.298705,0.195568)
    (0.344659,0.193931)
    (0.390614,0.193440)
    (0.436568,0.197975)
    (0.482523,0.200681)
    (0.528478,0.191961)
    (0.574432,0.198081)
    (0.620387,0.202910)
    (0.666341,0.198287)
    (0.712296,0.197386)
    (0.758250,0.201063)
    (0.804205,0.209615)
    (0.850160,0.196973)
    (0.896114,0.217705)
};
\addplot[
    color = {purple},
]
coordinates {
    (0.022977,0.175002)
    (0.068932,0.166065)
    (0.114886,0.183586)
    (0.160841,0.168230)
    (0.206796,0.187524)
    (0.252750,0.177597)
    (0.298705,0.194779)
    (0.344659,0.192107)
    (0.390614,0.191003)
    (0.436568,0.196449)
    (0.482523,0.197965)
    (0.528478,0.190033)
    (0.574432,0.194974)
    (0.620387,0.199753)
    (0.666341,0.194969)
    (0.712296,0.193414)
    (0.758250,0.192769)
    (0.804205,0.200320)
    (0.850160,0.186827)
    (0.896114,0.199920)
};
\addplot[
    color = {orange},
]
coordinates {
    (0.022977,0.175256)
    (0.068932,0.167370)
    (0.114886,0.184725)
    (0.160841,0.169363)
    (0.206796,0.187629)
    (0.252750,0.174762)
    (0.298705,0.192862)
    (0.344659,0.192689)
    (0.390614,0.190959)
    (0.436568,0.195733)
    (0.482523,0.201992)
    (0.528478,0.197611)
    (0.574432,0.200801)
    (0.620387,0.211897)
    (0.666341,0.210922)
    (0.712296,0.218004)
    (0.758250,0.220483)
    (0.804205,0.229869)
    (0.850160,0.214692)
    (0.896114,0.249886)
};

\nextgroupplot[
		y post scale=.4,
		ymax = 750,
		xlabel = {\small (a) Norm. Entropy},
]
\addplot[
    ybar,
    fill,
]
coordinates {
    (0.022977,159.000000)
    (0.068932,311.000000)
    (0.114886,424.000000)
    (0.160841,461.000000)
    (0.206796,503.000000)
    (0.252750,536.000000)
    (0.298705,551.000000)
    (0.344659,555.000000)
    (0.390614,578.000000)
    (0.436568,566.000000)
    (0.482523,537.000000)
    (0.528478,577.000000)
    (0.574432,561.000000)
    (0.620387,610.000000)
    (0.666341,641.000000)
    (0.712296,625.000000)
    (0.758250,641.000000)
    (0.804205,533.000000)
    (0.850160,469.000000)
    (0.896114,162.000000)
};

\end{groupplot}
\end{tikzpicture}\hspace{-2mm}%
\begin{tikzpicture}
\begin{groupplot}[
    group style = {group size=1 by 3,vertical sep=0pt},
		x tick label style={font=\tiny},
		y tick label style={font=\tiny},
		x post scale=.75
]
\nextgroupplot[
		legend pos = north east,
		legend style={font=\tiny},
		y post scale=.65
]
\addplot[
    color = {red},
]
coordinates {
    (0.024604,0.735592) +- (0.000000,0.192876)
    (0.073812,0.741860) +- (0.000000,0.188542)
    (0.123020,0.742153) +- (0.000000,0.188840)
    (0.172228,0.737160) +- (0.000000,0.184721)
    (0.221436,0.738736) +- (0.000000,0.188616)
    (0.270643,0.738467) +- (0.000000,0.188479)
    (0.319851,0.717412) +- (0.000000,0.186931)
    (0.369059,0.725522) +- (0.000000,0.184595)
    (0.418267,0.703477) +- (0.000000,0.185062)
    (0.467475,0.704822) +- (0.000000,0.176352)
    (0.516683,0.691206) +- (0.000000,0.183648)
    (0.565891,0.692188) +- (0.000000,0.179459)
    (0.615099,0.694770) +- (0.000000,0.173828)
    (0.664307,0.683508) +- (0.000000,0.173775)
    (0.713514,0.679484) +- (0.000000,0.171782)
    (0.762722,0.679774) +- (0.000000,0.179494)
    (0.811930,0.657024) +- (0.000000,0.185110)
    (0.861138,0.660238) +- (0.000000,0.187113)
    (0.910346,0.631884) +- (0.000000,0.198326)
    (0.959554,0.635190) +- (0.000000,0.162229)
};
\addlegendentry{MajorityVoting}
\addplot[
    color = {blue},
]
coordinates {
    (0.024604,0.761187) +- (0.000000,0.204453)
    (0.073812,0.761093) +- (0.000000,0.198926)
    (0.123020,0.758375) +- (0.000000,0.200067)
    (0.172228,0.754909) +- (0.000000,0.194654)
    (0.221436,0.749861) +- (0.000000,0.198243)
    (0.270643,0.747212) +- (0.000000,0.197359)
    (0.319851,0.722852) +- (0.000000,0.196932)
    (0.369059,0.728662) +- (0.000000,0.194821)
    (0.418267,0.707797) +- (0.000000,0.193647)
    (0.467475,0.708273) +- (0.000000,0.183360)
    (0.516683,0.692153) +- (0.000000,0.192598)
    (0.565891,0.692527) +- (0.000000,0.187868)
    (0.615099,0.695024) +- (0.000000,0.180627)
    (0.664307,0.682714) +- (0.000000,0.181613)
    (0.713514,0.679118) +- (0.000000,0.176936)
    (0.762722,0.678214) +- (0.000000,0.185381)
    (0.811930,0.657274) +- (0.000000,0.188757)
    (0.861138,0.659154) +- (0.000000,0.191136)
    (0.910346,0.633827) +- (0.000000,0.202198)
    (0.959554,0.633488) +- (0.000000,0.163216)
};
\addlegendentry{LTD-RBM}
\addplot[
    color = {green},
]
coordinates {
    (0.024604,0.769588) +- (0.000000,0.211266)
    (0.073812,0.774507) +- (0.000000,0.199586)
    (0.123020,0.768874) +- (0.000000,0.204323)
    (0.172228,0.761934) +- (0.000000,0.198628)
    (0.221436,0.752417) +- (0.000000,0.204208)
    (0.270643,0.751289) +- (0.000000,0.203331)
    (0.319851,0.724222) +- (0.000000,0.200985)
    (0.369059,0.728989) +- (0.000000,0.200037)
    (0.418267,0.707070) +- (0.000000,0.195447)
    (0.467475,0.706400) +- (0.000000,0.186527)
    (0.516683,0.690458) +- (0.000000,0.197498)
    (0.565891,0.689450) +- (0.000000,0.192420)
    (0.615099,0.692639) +- (0.000000,0.183644)
    (0.664307,0.678684) +- (0.000000,0.184529)
    (0.713514,0.676086) +- (0.000000,0.179129)
    (0.762722,0.677109) +- (0.000000,0.186296)
    (0.811930,0.656521) +- (0.000000,0.189081)
    (0.861138,0.658492) +- (0.000000,0.192815)
    (0.910346,0.635386) +- (0.000000,0.201477)
    (0.959554,0.633317) +- (0.000000,0.163499)
};
\addlegendentry{MLE}
\addplot[
    color = {purple},
]
coordinates {
    (0.024604,0.770829) +- (0.000000,0.202097)
    (0.073812,0.767040) +- (0.000000,0.198969)
    (0.123020,0.766086) +- (0.000000,0.197773)
    (0.172228,0.761308) +- (0.000000,0.194815)
    (0.221436,0.751005) +- (0.000000,0.200016)
    (0.270643,0.748471) +- (0.000000,0.199498)
    (0.319851,0.721828) +- (0.000000,0.197954)
    (0.369059,0.729360) +- (0.000000,0.195983)
    (0.418267,0.707253) +- (0.000000,0.193388)
    (0.467475,0.706611) +- (0.000000,0.184411)
    (0.516683,0.691993) +- (0.000000,0.194077)
    (0.565891,0.690913) +- (0.000000,0.189110)
    (0.615099,0.692915) +- (0.000000,0.182128)
    (0.664307,0.680227) +- (0.000000,0.182643)
    (0.713514,0.676950) +- (0.000000,0.177477)
    (0.762722,0.678183) +- (0.000000,0.185877)
    (0.811930,0.657251) +- (0.000000,0.187756)
    (0.861138,0.659484) +- (0.000000,0.192049)
    (0.910346,0.634420) +- (0.000000,0.200907)
    (0.959554,0.632973) +- (0.000000,0.163290)
};
\addlegendentry{LTM}
\addplot[
    color = {orange},
]
coordinates {
    (0.024604,0.727237) +- (0.000000,0.225685)
    (0.073812,0.713524) +- (0.000000,0.230050)
    (0.123020,0.723250) +- (0.000000,0.224945)
    (0.172228,0.711898) +- (0.000000,0.220538)
    (0.221436,0.702978) +- (0.000000,0.223474)
    (0.270643,0.706020) +- (0.000000,0.224784)
    (0.319851,0.671168) +- (0.000000,0.227421)
    (0.369059,0.690156) +- (0.000000,0.213975)
    (0.418267,0.672117) +- (0.000000,0.206948)
    (0.467475,0.667557) +- (0.000000,0.202464)
    (0.516683,0.659408) +- (0.000000,0.206809)
    (0.565891,0.661722) +- (0.000000,0.203879)
    (0.615099,0.667181) +- (0.000000,0.188916)
    (0.664307,0.661994) +- (0.000000,0.188044)
    (0.713514,0.660010) +- (0.000000,0.184996)
    (0.762722,0.664749) +- (0.000000,0.188242)
    (0.811930,0.648075) +- (0.000000,0.191604)
    (0.861138,0.657482) +- (0.000000,0.191786)
    (0.910346,0.631579) +- (0.000000,0.200408)
    (0.959554,0.631595) +- (0.000000,0.165997)
};
\addlegendentry{2-Estimates}

\nextgroupplot[
	y post scale=.65
]
\addplot[
    color = {red},
]
coordinates {
    (0.024604,0.192876)
    (0.073812,0.188542)
    (0.123020,0.188840)
    (0.172228,0.184721)
    (0.221436,0.188616)
    (0.270643,0.188479)
    (0.319851,0.186931)
    (0.369059,0.184595)
    (0.418267,0.185062)
    (0.467475,0.176352)
    (0.516683,0.183648)
    (0.565891,0.179459)
    (0.615099,0.173828)
    (0.664307,0.173775)
    (0.713514,0.171782)
    (0.762722,0.179494)
    (0.811930,0.185110)
    (0.861138,0.187113)
    (0.910346,0.198326)
    (0.959554,0.162229)
};
\addplot[
    color = {blue},
]
coordinates {
    (0.024604,0.204453)
    (0.073812,0.198926)
    (0.123020,0.200067)
    (0.172228,0.194654)
    (0.221436,0.198243)
    (0.270643,0.197359)
    (0.319851,0.196932)
    (0.369059,0.194821)
    (0.418267,0.193647)
    (0.467475,0.183360)
    (0.516683,0.192598)
    (0.565891,0.187868)
    (0.615099,0.180627)
    (0.664307,0.181613)
    (0.713514,0.176936)
    (0.762722,0.185381)
    (0.811930,0.188757)
    (0.861138,0.191136)
    (0.910346,0.202198)
    (0.959554,0.163216)
};
\addplot[
    color = {green},
]
coordinates {
    (0.024604,0.211266)
    (0.073812,0.199586)
    (0.123020,0.204323)
    (0.172228,0.198628)
    (0.221436,0.204208)
    (0.270643,0.203331)
    (0.319851,0.200985)
    (0.369059,0.200037)
    (0.418267,0.195447)
    (0.467475,0.186527)
    (0.516683,0.197498)
    (0.565891,0.192420)
    (0.615099,0.183644)
    (0.664307,0.184529)
    (0.713514,0.179129)
    (0.762722,0.186296)
    (0.811930,0.189081)
    (0.861138,0.192815)
    (0.910346,0.201477)
    (0.959554,0.163499)
};
\addplot[
    color = {purple},
]
coordinates {
    (0.024604,0.202097)
    (0.073812,0.198969)
    (0.123020,0.197773)
    (0.172228,0.194815)
    (0.221436,0.200016)
    (0.270643,0.199498)
    (0.319851,0.197954)
    (0.369059,0.195983)
    (0.418267,0.193388)
    (0.467475,0.184411)
    (0.516683,0.194077)
    (0.565891,0.189110)
    (0.615099,0.182128)
    (0.664307,0.182643)
    (0.713514,0.177477)
    (0.762722,0.185877)
    (0.811930,0.187756)
    (0.861138,0.192049)
    (0.910346,0.200907)
    (0.959554,0.163290)
};
\addplot[
    color = {orange},
]
coordinates {
    (0.024604,0.225685)
    (0.073812,0.230050)
    (0.123020,0.224945)
    (0.172228,0.220538)
    (0.221436,0.223474)
    (0.270643,0.224784)
    (0.319851,0.227421)
    (0.369059,0.213975)
    (0.418267,0.206948)
    (0.467475,0.202464)
    (0.516683,0.206809)
    (0.565891,0.203879)
    (0.615099,0.188916)
    (0.664307,0.188044)
    (0.713514,0.184996)
    (0.762722,0.188242)
    (0.811930,0.191604)
    (0.861138,0.191786)
    (0.910346,0.200408)
    (0.959554,0.165997)
};

\nextgroupplot[
	y post scale=.4,
	ymax = 1150,
	xlabel = {\small (b) Frequency of copied claims},
]
\addplot[
    ybar,
    fill,
]
coordinates {
    (0.024604,977.000000)
    (0.073812,393.000000)
    (0.123020,470.000000)
    (0.172228,489.000000)
    (0.221436,470.000000)
    (0.270643,525.000000)
    (0.319851,575.000000)
    (0.369059,555.000000)
    (0.418267,616.000000)
    (0.467475,594.000000)
    (0.516683,597.000000)
    (0.565891,643.000000)
    (0.615099,643.000000)
    (0.664307,594.000000)
    (0.713514,557.000000)
    (0.762722,488.000000)
    (0.811930,376.000000)
    (0.861138,245.000000)
    (0.910346,140.000000)
    (0.959554,53.000000)
};

\end{groupplot}
\end{tikzpicture}\hspace{-2mm}%
\begin{tikzpicture}
\begin{groupplot}[
    group style = {group size=1 by 3,vertical sep=0pt},
		yticklabel style={
        /pgf/number format/fixed,
		},
		x tick label style={font=\tiny},
		y tick label style={font=\tiny},
		x post scale=.75
]
\nextgroupplot[
		legend pos = south east,
		legend style={font=\tiny},
		y post scale=.65,
]
\addplot[
    color = {red},
]
coordinates {
    (0.094595,0.130903) +- (0.000000,0.089665)
    (0.140476,0.714602) +- (0.000000,0.000000)
    (0.186357,0.297177) +- (0.000000,0.000000)
    (0.232238,0.248793) +- (0.000000,0.063319)
    (0.278119,0.325090) +- (0.000000,0.099586)
    (0.324000,0.355889) +- (0.000000,0.077038)
    (0.369880,0.384772) +- (0.000000,0.077053)
    (0.415761,0.440049) +- (0.000000,0.081846)
    (0.461642,0.493255) +- (0.000000,0.078009)
    (0.507523,0.557961) +- (0.000000,0.085564)
    (0.553404,0.610748) +- (0.000000,0.087224)
    (0.599285,0.665600) +- (0.000000,0.091908)
    (0.645165,0.723707) +- (0.000000,0.092919)
    (0.691046,0.772375) +- (0.000000,0.093566)
    (0.736927,0.816572) +- (0.000000,0.085751)
    (0.782808,0.856924) +- (0.000000,0.074371)
    (0.828689,0.892320) +- (0.000000,0.074142)
    (0.874569,0.923477) +- (0.000000,0.057971)
    (0.920450,0.944685) +- (0.000000,0.047315)
    (0.966331,0.962389) +- (0.000000,0.027258)
};
\addlegendentry{MajorityVoting}
\addplot[
    color = {blue},
]
coordinates {
    (0.094595,0.130927) +- (0.000000,0.092619)
    (0.140476,0.694690) +- (0.000000,0.000000)
    (0.186357,0.295437) +- (0.000000,0.000000)
    (0.232238,0.233688) +- (0.000000,0.069691)
    (0.278119,0.316974) +- (0.000000,0.100878)
    (0.324000,0.346627) +- (0.000000,0.082708)
    (0.369880,0.371146) +- (0.000000,0.086414)
    (0.415761,0.428639) +- (0.000000,0.094927)
    (0.461642,0.488992) +- (0.000000,0.094755)
    (0.507523,0.563426) +- (0.000000,0.104532)
    (0.553404,0.622673) +- (0.000000,0.100255)
    (0.599285,0.675909) +- (0.000000,0.107317)
    (0.645165,0.736374) +- (0.000000,0.106928)
    (0.691046,0.784940) +- (0.000000,0.101671)
    (0.736927,0.828604) +- (0.000000,0.093056)
    (0.782808,0.868008) +- (0.000000,0.078732)
    (0.828689,0.901787) +- (0.000000,0.075679)
    (0.874569,0.930600) +- (0.000000,0.058009)
    (0.920450,0.949433) +- (0.000000,0.047677)
    (0.966331,0.966038) +- (0.000000,0.028033)
};
\addlegendentry{LTD-RBM}
\addplot[
    color = {green},
]
coordinates {
    (0.094595,0.137302) +- (0.000000,0.087909)
    (0.140476,0.752212) +- (0.000000,0.000000)
    (0.186357,0.293503) +- (0.000000,0.000000)
    (0.232238,0.236399) +- (0.000000,0.087036)
    (0.278119,0.316106) +- (0.000000,0.102374)
    (0.324000,0.341527) +- (0.000000,0.091773)
    (0.369880,0.362179) +- (0.000000,0.096557)
    (0.415761,0.425208) +- (0.000000,0.115522)
    (0.461642,0.490320) +- (0.000000,0.114236)
    (0.507523,0.568586) +- (0.000000,0.119395)
    (0.553404,0.628100) +- (0.000000,0.113554)
    (0.599285,0.680899) +- (0.000000,0.116989)
    (0.645165,0.739424) +- (0.000000,0.112216)
    (0.691046,0.786540) +- (0.000000,0.106362)
    (0.736927,0.829435) +- (0.000000,0.095959)
    (0.782808,0.867765) +- (0.000000,0.080353)
    (0.828689,0.902267) +- (0.000000,0.076189)
    (0.874569,0.931159) +- (0.000000,0.057388)
    (0.920450,0.949194) +- (0.000000,0.047741)
    (0.966331,0.965325) +- (0.000000,0.027979)
};
\addlegendentry{MLE}
\addplot[
    color = {purple},
]
coordinates {
    (0.094595,0.137859) +- (0.000000,0.088473)
    (0.140476,0.727876) +- (0.000000,0.000000)
    (0.186357,0.295953) +- (0.000000,0.000000)
    (0.232238,0.229666) +- (0.000000,0.074969)
    (0.278119,0.309020) +- (0.000000,0.103102)
    (0.324000,0.349003) +- (0.000000,0.087092)
    (0.369880,0.371962) +- (0.000000,0.095154)
    (0.415761,0.429455) +- (0.000000,0.106661)
    (0.461642,0.494771) +- (0.000000,0.110550)
    (0.507523,0.570106) +- (0.000000,0.111170)
    (0.553404,0.625304) +- (0.000000,0.105933)
    (0.599285,0.677392) +- (0.000000,0.110802)
    (0.645165,0.736397) +- (0.000000,0.110017)
    (0.691046,0.784523) +- (0.000000,0.104497)
    (0.736927,0.828856) +- (0.000000,0.094314)
    (0.782808,0.867662) +- (0.000000,0.079202)
    (0.828689,0.901600) +- (0.000000,0.076108)
    (0.874569,0.929882) +- (0.000000,0.057131)
    (0.920450,0.948627) +- (0.000000,0.047525)
    (0.966331,0.964784) +- (0.000000,0.026839)
};
\addlegendentry{LTM}
\addplot[
    color = {orange},
]
coordinates {
    (0.094595,0.142768) +- (0.000000,0.083913)
    (0.140476,0.756637) +- (0.000000,0.000000)
    (0.186357,0.294664) +- (0.000000,0.000000)
    (0.232238,0.221855) +- (0.000000,0.077571)
    (0.278119,0.320620) +- (0.000000,0.102525)
    (0.324000,0.351038) +- (0.000000,0.093556)
    (0.369880,0.364562) +- (0.000000,0.106078)
    (0.415761,0.415410) +- (0.000000,0.127526)
    (0.461642,0.475583) +- (0.000000,0.130380)
    (0.507523,0.535376) +- (0.000000,0.139928)
    (0.553404,0.588316) +- (0.000000,0.134508)
    (0.599285,0.632677) +- (0.000000,0.146547)
    (0.645165,0.690124) +- (0.000000,0.144013)
    (0.691046,0.739117) +- (0.000000,0.141208)
    (0.736927,0.777010) +- (0.000000,0.149292)
    (0.782808,0.832454) +- (0.000000,0.117911)
    (0.828689,0.871081) +- (0.000000,0.114723)
    (0.874569,0.908140) +- (0.000000,0.093680)
    (0.920450,0.935870) +- (0.000000,0.070986)
    (0.966331,0.960430) +- (0.000000,0.031214)
};
\addlegendentry{2-Estimates}
\addplot[
    color = {black},
		dashed
]
coordinates {
    (0.094595,0.094595)
    (0.966331,0.966331)
};
\nextgroupplot[
		y post scale=.65
]
\addplot[
    color = {red},
]
coordinates {
    (0.094595,0.089665)
    (0.140476,0.000000)
    (0.186357,0.000000)
    (0.232238,0.063319)
    (0.278119,0.099586)
    (0.324000,0.077038)
    (0.369880,0.077053)
    (0.415761,0.081846)
    (0.461642,0.078009)
    (0.507523,0.085564)
    (0.553404,0.087224)
    (0.599285,0.091908)
    (0.645165,0.092919)
    (0.691046,0.093566)
    (0.736927,0.085751)
    (0.782808,0.074371)
    (0.828689,0.074142)
    (0.874569,0.057971)
    (0.920450,0.047315)
    (0.966331,0.027258)
};
\addplot[
    color = {blue},
]
coordinates {
    (0.094595,0.092619)
    (0.140476,0.000000)
    (0.186357,0.000000)
    (0.232238,0.069691)
    (0.278119,0.100878)
    (0.324000,0.082708)
    (0.369880,0.086414)
    (0.415761,0.094927)
    (0.461642,0.094755)
    (0.507523,0.104532)
    (0.553404,0.100255)
    (0.599285,0.107317)
    (0.645165,0.106928)
    (0.691046,0.101671)
    (0.736927,0.093056)
    (0.782808,0.078732)
    (0.828689,0.075679)
    (0.874569,0.058009)
    (0.920450,0.047677)
    (0.966331,0.028033)
};
\addplot[
    color = {green},
]
coordinates {
    (0.094595,0.087909)
    (0.140476,0.000000)
    (0.186357,0.000000)
    (0.232238,0.087036)
    (0.278119,0.102374)
    (0.324000,0.091773)
    (0.369880,0.096557)
    (0.415761,0.115522)
    (0.461642,0.114236)
    (0.507523,0.119395)
    (0.553404,0.113554)
    (0.599285,0.116989)
    (0.645165,0.112216)
    (0.691046,0.106362)
    (0.736927,0.095959)
    (0.782808,0.080353)
    (0.828689,0.076189)
    (0.874569,0.057388)
    (0.920450,0.047741)
    (0.966331,0.027979)
};
\addplot[
    color = {purple},
]
coordinates {
    (0.094595,0.088473)
    (0.140476,0.000000)
    (0.186357,0.000000)
    (0.232238,0.074969)
    (0.278119,0.103102)
    (0.324000,0.087092)
    (0.369880,0.095154)
    (0.415761,0.106661)
    (0.461642,0.110550)
    (0.507523,0.111170)
    (0.553404,0.105933)
    (0.599285,0.110802)
    (0.645165,0.110017)
    (0.691046,0.104497)
    (0.736927,0.094314)
    (0.782808,0.079202)
    (0.828689,0.076108)
    (0.874569,0.057131)
    (0.920450,0.047525)
    (0.966331,0.026839)
};
\addplot[
    color = {orange},
]
coordinates {
    (0.094595,0.083913)
    (0.140476,0.000000)
    (0.186357,0.000000)
    (0.232238,0.077571)
    (0.278119,0.102525)
    (0.324000,0.093556)
    (0.369880,0.106078)
    (0.415761,0.127526)
    (0.461642,0.130380)
    (0.507523,0.139928)
    (0.553404,0.134508)
    (0.599285,0.146547)
    (0.645165,0.144013)
    (0.691046,0.141208)
    (0.736927,0.149292)
    (0.782808,0.117911)
    (0.828689,0.114723)
    (0.874569,0.093680)
    (0.920450,0.070986)
    (0.966331,0.031214)
};

\nextgroupplot[
	y post scale=.4,
	ymax = 1100,
	xlabel = {\small (c) Avg. source accuracy},
]
\addplot[
    ybar,
    fill,
]
coordinates {
    (0.094595,4.000000)
    (0.140476,1.000000)
    (0.186357,1.000000)
    (0.232238,11.000000)
    (0.278119,27.000000)
    (0.324000,95.000000)
    (0.369880,281.000000)
    (0.415761,585.000000)
    (0.461642,789.000000)
    (0.507523,853.000000)
    (0.553404,870.000000)
    (0.599285,889.000000)
    (0.645165,933.000000)
    (0.691046,902.000000)
    (0.736927,904.000000)
    (0.782808,861.000000)
    (0.828689,838.000000)
    (0.874569,763.000000)
    (0.920450,333.000000)
    (0.966331,60.000000)
};

\end{groupplot}
\end{tikzpicture}%
\caption{ Average accuracy in relation to (a) the entropy as a measure for ambiguousness (b) the frequency of copies as a measure for source dependencies and (c) the quality of sources. Each plot shows the average accuracy (top), its standard deviation (middle) and the number of datasets that were used to compute the above values (bottom).}%
\label{fig:acc_vs_characteristic}%
\end{figure*}
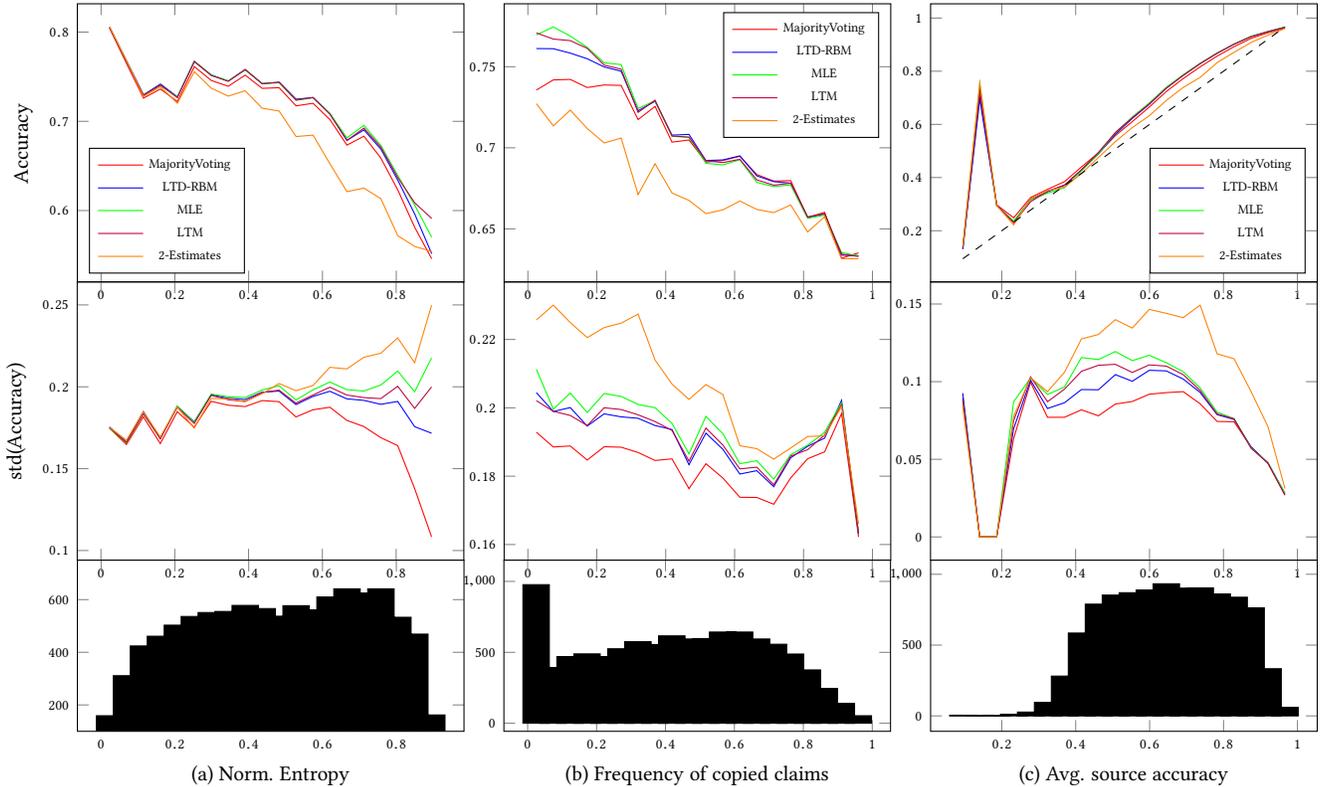

\subsection{Synthetic Data}
\label{sec:eval_synthetic}
In order to evaluate how well the algorithms perform on different types of datasets, we randomly generated \dsCount{} different synthetic datasets with different characteristics. 
Using these datasets, we can evaluate the influence of multiple aspects on the truth discovery. 
In the following, we first briefly describe the generation of our synthetic data and afterwards present our results on these datasets.

\paragraph{Generating Synthetic Data}
In section \ref{sec:algorithm}, we described the assumptions that were made for LTD-RBM: given the latent truth, we assumed independence of sources as well as independence of facts. Normally, real-world scenarios do not fulfill these assumptions. Hence, we also generate the synthetic datasets in a way that break these assumptions: we create categorical data as described in \ref{sec:data_model} and include copying sources. 

In the following, we describe our dataset generation process. For each dataset we randomly choose values for a set of parameters and based on these parameters, we randomly generate a set of source models. Based on the source models and the dataset parameters, we then generate the dataset.
This random generation allows us to cover a huge variability of dataset characteristics without the need to create each possible combination.

The dataset parameters are the number of sources, the number of entity attributes, the expected claim frequency of sources, the expected average source accuracy and the probability for a source to be a copying source.
For the source frequency and accuracy, we also choose a value of variability. 
Combined with the expected value, we get Beta-Distributions by using the variability value as sum of the two distribution parameters. 

Based on these dataset parameters, we can randomly choose the following source model parameters: The claim frequency and the accuracy are drawn from the above mentioned Beta-Distributions. With the above mentioned probability, a source is a copying source. In this case, a random source is assigned as source for the copies and a random frequency of copies is assigned. This frequency models how many claims are original claims and how many are copied from the other source.

Based on this model, we randomly generated $\dsCount$ datasets of variable size from 200 up to 500.000 claims.
In order to evaluate the influence of certain aspects on the LTD performance, we grouped the datasets according to that aspect and compute for each group the average accuracy and the standard deviation of the accuracy. This allows us to analyze the influence of the aspect on the performance of different LTD-Algorithms.

In the following, we will highlight some aspects. Note that due to the randomness of our datasets, some results are based on a small number of datasets. This can lead to noise and outliers in the diagrams. We will always show the number of datasets that lead to a result to indicate where such artifacts can occur.

\paragraph{Ambiguousness} For some facts or entity attributes, most sources favor the same value, while for other facts the sources are more ambiguous in their statements. These characteristics can be extended on datasets. We use the normalized entropy per fact / entity attribute as measure for the ambiguousness that goes from $0$ (all sources always agree) to $1$ (all possible values receive support from the same number of sources). Figure~\ref{fig:acc_vs_characteristic} (a) shows how the different algorithms perform for different levels of ambiguousness.

As expected, all algorithms have problems with high entropy up to the very ambiguous point where all answers receive the same number of supporting sources. It can be seen that LTD-RBM is among the best algorithms except for the border cases where the low number of samples leads to sampling noise with less significant results.
Furthermore, LTD-algorithms differ in their ability to produce stable results for datasets with high entropy. Here, the variance of LTD-RBM shows that its effectiveness is less dataset-dependent than for the other state-of-the-art algorithms.

\paragraph{Source Dependencies} As previously mentioned, we intentionally break the assumption that all sources are independent by allowing copying sources that partially copy from other sources. This allows us to analyze how well the algorithms perform when the independence assumption is broken. Figure~\ref{fig:acc_vs_characteristic} (b) shows how the different algorithms perform for different levels of dependence.

Similarly to the ambiguousness results, it can be seen that the accuracy gets lower with increasing frequency of copying. The reason is that all algorithms expect independent sources. Again, LTD-RBM is one of the topmost algorithms that also provides more stable results than the other state-of-the-art algorithms.

\paragraph{Source Qualities} One major influence for the outcome of the LTD is the quality of the sources. It is much easier to infer the hidden truth when most sources make most of the time true claims. We take the average accuracy of sources as a measurement for the source quality in a dataset and measure in this way how well the algorithms deal with different levels of source quality. The results can be seen in Fig.~\ref{fig:acc_vs_characteristic} (c).
It can be seen that we only generated a few outlier datasets with an average source accuracy below $0.3$. This leads to results that are heavily affected by noise. We plotted these datasets for the sake of completeness, since they are also included in the data for the other plots. Again, LTD-RBM shows top effectiveness while providing the most stable results among the state-of-the-art competitors.
The plot shows that all algorithms are sometimes just slightly better than the average source. The reason is that we set up the dataset generation process to include hard problems, e.g. by including many copying sources. The aggregation over all datasets then results in the shown plot.

\paragraph{Robustness} As our experiments show, LTD-RBM produces stable results. This can be seen in the computed standard deviations. Only Majority Voting has a lower variance, but shows at the same time the lowest effectiveness. This means that for unknown datasets, the results of our method are more stable and predictable than for the other state-of the-art competitors.
We hypothesize that one reason for the robustness of LTD-RBM is that it updates the weights on mini-batches, while the other methods update globally over all claims.

\paragraph{Computational costs} In difference to the previous aspects, the computational costs do not deal with the accuracy, but with the runtime of an algorithm. We measured the runtime in relation to the number of claims in a dataset. Figure~\ref{fig:time_vs_claims} shows that the runtime of LTD-RBM is only outperformed by the simple Majority Voting.

\subsection{Real World Data}
\label{sec:eval_real}
Additionally to our experiments on synthetic datasets, we also evaluate the overall performance on real-world datasets.
For this, we use two different publicly available datasets\footnote{Both were used in~\cite{Waguih2014} and can be found at: \url{http://da.qcri.org/dafna}.}: A flight dataset and a weather dataset.
We found that the flight and the weather data contained raw values that cause problems for the LTD algorithms. The reason is that all algorithms work with categorical values. Thus, technically identical values with different representations are considered as completely different values. An example are dates that come in different formats from different sources. In order to compensate for this effect, we applied a preprocessing step that normalized the representations. As a result, we got four datasets for our tests (two datasets, each with and without preprocessing).

Table~\ref{tab:dataset_characteristics} gives an overview of these datasets. The raw flights dataset contains dates in a large variety of formats. That makes it nearly impossible to find the correct value in a categorical sense. For this purpose, we normalized the date strings and obtained much better results with all algorithms. 

\begin{table}%
\small
\begin{tabular}{l|rrrr}
\hline\hline 
\multicolumn{5}{c}{\cellcolor{gray!25}LTD-RBM}\\ 
\hline 
Optimized For & Flight (0) & Flight (1) & Weather (0) & Weather (1)\\ 
\hline 
Flight (0) & 11.46 \% & 64.72 \% & 58.71 \% & 70.86 \%\\ 
Flight (1) & 9.11 \% & 76.61 \% & 55.97 \% & 58.09 \%\\ 
Weather (0) & 8.16 \% & 76.28 \% & 71.56 \% & 67.68 \%\\ 
Weather (1) & 10.99 \% & 64.75 \% & 61.42 \% & 72.17 \%\\ 
\hline 
Mean & \colCol{30} 9.58 \% & \colCol{100} 71.72 \% & \colCol{100} 63.79 \% & \colCol{100} 66.58 \%\\ 
Std & \colCol{58} 1.40 \% & \colCol{100} 5.70 \% & \colCol{91} 6.47 \% & \colCol{93} 5.08 \%\\ 
Max & \colCol{26} 11.46 \% & \colCol{100} 76.61 \% & \colCol{69} 71.56 \% & \colCol{100} 72.17 \%\\ 
\hline\hline 
\multicolumn{5}{c}{}\\[-2mm] 
\hline\hline 
\multicolumn{5}{c}{\cellcolor{gray!25}LTD-RBM-C}\\ 
\hline 
Optimized For & Flight (0) & Flight (1) & Weather (0) & Weather (1)\\ 
\hline 
Flight (0) & 12.81 \% & 53.50 \% & 8.05 \% & 10.10 \%\\ 
Flight (1) & 9.39 \% & 76.55 \% & 57.58 \% & 56.72 \%\\ 
Weather (0) & 11.30 \% & 76.19 \% & 73.41 \% & 72.07 \%\\ 
Weather (1) & 9.34 \% & 76.14 \% & 73.33 \% & 71.81 \%\\ 
\hline 
Mean & \colCol{100} 10.43 \% & \colCol{100} 71.76 \% & \colCol{69} 54.81 \% & \colCol{55} 54.37 \%\\ 
Std & \colCol{57} 1.41 \% & \colCol{60} 9.13 \% & \colCol{22} 24.21 \% & \colCol{9} 22.93 \%\\ 
Max & \colCol{96} 12.81 \% & \colCol{89} 76.55 \% & \colCol{100} 73.41 \% & \colCol{99} 72.07 \%\\ 
\hline\hline 
\multicolumn{5}{c}{}\\[-2mm] 
\hline\hline 
\multicolumn{5}{c}{\cellcolor{gray!25}MLE}\\ 
\hline 
Optimized For & Flight (0) & Flight (1) & Weather (0) & Weather (1)\\ 
\hline 
Flight (0) & 10.97 \% & 74.21 \% & 57.72 \% & 57.41 \%\\ 
Flight (1) & 9.26 \% & 76.07 \% & 61.86 \% & 60.82 \%\\ 
Weather (0) & 9.27 \% & 46.14 \% & 68.36 \% & 66.53 \%\\ 
Weather (1) & 9.27 \% & 46.14 \% & 68.36 \% & 66.53 \%\\ 
\hline 
Mean & \colCol{33} 9.61 \% & \colCol{0} 63.72 \% & \colCol{100} 63.75 \% & \colCol{85} 62.50 \%\\ 
Std & \colCol{100} 0.68 \% & \colCol{0} 14.37 \% & \colCol{100} 4.10 \% & \colCol{100} 3.55 \%\\ 
Max & \colCol{0} 10.97 \% & \colCol{0} 76.07 \% & \colCol{16} 68.36 \% & \colCol{24} 66.53 \%\\ 
\hline\hline 
\multicolumn{5}{c}{}\\[-2mm] 
\hline\hline 
\multicolumn{5}{c}{\cellcolor{gray!25}LTM}\\ 
\hline 
Optimized For & Flight (0) & Flight (1) & Weather (0) & Weather (1)\\ 
\hline 
Flight (0) & 12.89 \% & 47.22 \% & 38.60 \% & 48.08 \%\\ 
Flight (1) & 6.76 \% & 76.38 \% & 0.62 \% & 9.69 \%\\ 
Weather (0) & 10.83 \% & 76.28 \% & 66.72 \% & 63.95 \%\\ 
Weather (1) & 6.77 \% & 65.70 \% & 67.42 \% & 64.77 \%\\ 
\hline 
Mean & \colCol{0} 9.20 \% & \colCol{32} 66.26 \% & \colCol{0} 34.78 \% & \colCol{0} 39.29 \%\\ 
Std & \colCol{0} 2.38 \% & \colCol{43} 10.64 \% & \colCol{0} 29.79 \% & \colCol{0} 24.79 \%\\ 
Max & \colCol{100} 12.89 \% & \colCol{57} 76.38 \% & \colCol{0} 67.42 \% & \colCol{0} 64.77 \%\\ 
\hline\hline 
\multicolumn{5}{c}{}\\[-2mm] 
\end{tabular}

\vspace{1mm}
\caption{\textbf{Optimization dependency}. These tables show the influence of the hyperparameters on the effectiveness of the algorithms. We get different reasonable hyperparameter settings by optimizing the settings for specific datasets; these optimized settings for one dataset are then applied to the other datasets. Hence, naturally the best accuracy results are observed in the diagonal of the upper parts.
}
\label{tab:opt_cross}
\vspace{-5mm}
\end{table}

We applied LTD-RBM, MLE and LTM to the 4 datasets. In addition, we added the categorical version of our algorithm (LTD-RBM-C) to see if there are benefits from the modification. Beside the overall effectiveness (in terms of accuracy), we also focused on the robustness of the methods, especially the robustness towards the choice of hyperparameters. For an unknown dataset, it is often difficult to determine good hyperparameters. Thus, it is desirable to have an algorithm that is not strongly influenced by the choice of the hyperparameters.

In order to evaluate the robustness, we individually optimized the hyperparameters of each algorithm for each dataset. This gave us for each dataset an optimal parameter combination. Afterwards, we applied all combinations to all datasets. This allows us not only to measure the overall performance, but also to see how well the algorithms perform with poorly chosen hyperparameters.

We used an alternating scheme~\cite{Franek2011alternating} for the optimization to automatically find a local optimum hyperparameter combination for each algorithm. This scheme iteratively optimizes one parameter by fixing all other parameters until a local optimum is reached. This optimization was done on the prior source reliabilities, the learning rate (with decay) and the 2-Estimates-specific parameter.
We used the recognition accuracy as optimization criterion. 

The results of our experiment can be seen in Tab.~\ref{tab:opt_cross}.
It can be seen that LTD-RBM is the best choice for the preprocessed datasets. Furthermore, it provides stable results over multiple hyperparameter settings, which makes it a very robust algorithm. As for the categorical version: it is especially strong with unprocessed datasets, where there are various different claimed values for each attribute. Normalization steps in the preprocessing reduce this number - and thus the effect of the build-in constraint of only one true value. These benefits come with the drawback of lower robustness compared to the original RBM algorithm.

Comparing our results with previous works, it can be seen that the results do not always match. This has multiple reasons: our experiments show that preprocessing has a strong impact, but is often not described in detail. Furthermore, the hyperparameters have an influence (depending on how robust the algorithm is). As a last point: at least in our version of the datasets, there were conflicts in the ground truth data. We solved this by removing all conflicting entries. To the best of our knowledge, these conflicts were not dealt with in prior work.

\begin{table}[bt]
\centering
\begin{tabular}{l|rrrr}
DataSet & F. & F. Prep. & W. & W. Prep.\\
\hline
\#Attr. & 207,908 & 207,908 & 30,317 & 30,181\\
\#Claims & 2,849,984 & 2,844,030 & 307,335 & 306,034\\
\#Sources & 38 & 38 & 16 & 16\\
Entropy  & 0.8792 & 0.761 & 0.5561 & 0.4809\\
Avg-Acc & 6.15\%  & 45.69\% & 35.59\% & 40.93\% \\
Ground truth 
&7.71\%&7.71\%&62.96\%&63.25\%\\
\hline
\end{tabular}
\vspace{1mm}
\caption{Characterization of real-world datasets.}
\label{tab:dataset_characteristics}
\vspace{-7mm}
\end{table}

\begin{figure}%
\begin{tikzpicture}
\begin{axis}[
    legend style = {at={(0.95,0.65)},anchor=east,font=\small},
    ylabel = {Accuracy},
    xlabel = {initial false positive rate},
		y post scale=.6,
]
\addplot[
    mark=square,
    color=blue,
    ]
    coordinates {
(0.050000,0.765677)(0.100000,0.766135)(0.150000,0.766135)(0.200000,0.766155)(0.250000,0.766010)(0.300000,0.765760)(0.350000,0.765449)(0.400000,0.765220)(0.450000,0.764825)(0.500000,0.764867)(0.550000,0.764846)(0.600000,0.764430)(0.650000,0.764347)(0.700000,0.764326)(0.750000,0.764264)
    };
\addlegendentry{RBM}
\addplot[
    mark=square,
    color=green,
    ]
    coordinates {
(0.050000,0.244653)(0.100000,0.237451)(0.150000,0.409678)(0.200000,0.678119)(0.250000,0.760554)(0.300000,0.760678)(0.350000,0.760491)(0.400000,0.760616)(0.450000,0.566378)(0.500000,0.306334)(0.550000,0.061358)(0.600000,0.061616)(0.650000,0.063692)(0.700000,0.066871)(0.750000,0.067649)
    };
\addlegendentry{MLE}
\addplot[
    mark=square,
    color=purple,
    ]
    coordinates {
(0.050000,0.762456)(0.100000,0.762390)(0.150000,0.762698)(0.200000,0.763004)(0.250000,0.764157)(0.300000,0.762911)(0.350000,0.541673)(0.400000,0.068797)(0.450000,0.071455)(0.500000,0.071112)(0.550000,0.067750)(0.600000,0.068080)(0.650000,0.069961)(0.700000,0.070743)(0.750000,0.071188)
    };
\addlegendentry{LTM}
\end{axis}
\end{tikzpicture}
\caption{Example for the robustness towards the choice of hyperparameters. It shows how much the initial fpr influences the accuracy on the preprocessed flights dataset.}%
\label{fig:param_sensitivity}%
\vspace{-5mm}
\end{figure}
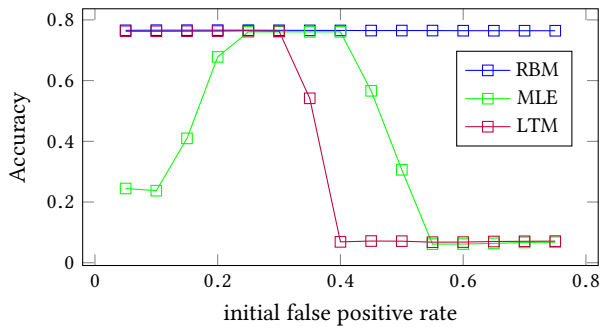

Besides that, we also took the optimal choice of hyperparameters and modified each hyperparameter. Figure~\ref{fig:param_sensitivity} shows an example for how sensitive the LTD-algorithms react to changes in their hyperparameters. Note that the diagram does not include 2-Estimates, since the shown parameters are only shared by the other three algorithms.
It can be seen that LTD-RBM works independently well for all false positive rates. In comparison, LTM and MLE are strongly affected by the choice of initial / prior false positive rate. MLE even performs poorly for both, too high as well as too low values. This makes it difficult to find a good value.

\section{Conclusion}
\label{sec:conclusion}
We have introduced an RBM-based approach to latent truth discovery. In various experiments on multiple datasets, we compared LTD-RBM to state-of-the-art LTD approaches. LTD-RBM shows a highly competitive performance in all conducted experiments and in terms of an overall consideration of effectiveness, efficiency and robustness, LTD-RBM outperforms all of its competitors. Especially, in terms of robustness, varying data quality and varying dataset properties (such as underlying source accuracy, amount of copied claims or difficulty of the truth discovery process in terms of normalized entropy), LTD-RBM shows the desired behavior.

Given the generality of the model behind LTD-RBM, we are planning to investigate the capability of LTD-RBM to deal with continuous values and predictions. An interesting challenge is to expand the LTD-RBM model to measure the confidence of predictions in an online fashion. The extension of the LTD-RBM model to deal with active learning scenarios is a further point on our agenda. In any case, we are confident that LTD-RBM will be the method of choice for many LTD tasks on large datasets.

\bibliographystyle{ACM-Reference-Format}
\bibliography{references}
\appendix
\section{Computing the hidden Bias}
\label{app:bias}
\begin{align*}
		&\sigma(b+\sum_sv_sw_s) 
	  \stackrel{(\ref{eqn:prob_hidden})}= P(h=1|\vec{v})
		=\frac{P(\vec{v}|h=1) P(h=1)}{P(\vec{v})}\\
		={}&\frac{P(\vec{v}|h=1) P(h=1)}{P(\vec{v}|h=1) P(h=1) + P(\vec{v}|h=0) P(h=0)}\\
		={}&\frac{1}{1 + \frac{P(\vec{v}|h=0) (1-p_T)}{P(\vec{v}|h=1) p_T}}\\
	\Rightarrow {} & b + \sum_sv_sw_s 
		= -\log\left(\frac{P(\vec{v}|h=0) (1-p_T)}{P(\vec{v}|h=1) p_T}\right)\\
		={}&-\log\left(\frac{(1-p_T)\prod_sP(v_s|h=0)}{p_T\prod_sP(v_s|h=1)}\right)\\
		\stackrel{(\ref{eqn:rbm_tpr},\ref{eqn:rbm_fpr})}={}&-\log\left(\frac{(1-p_T)}{p_T}\right) - \sum_s\log\left(\frac{\fpr_s^{v_s}(1-\fpr_s)^{1-v_s}}{\tpr_s^{v_s}(1-\tpr_s)^{1-v_s}}\right)\\
		={}&\log(p_T) - \log(1-p_T)  
		\\&- \sum_s\left(v_s\log(\fpr_s) + (1-v_s)\log(1-\fpr_s)\right.
		\\& \hphantom{\sum_s}- \left. v_s\log(\tpr_s)-(1-v_s)\log(1-\tpr_s)\right)\\
		={}&\log(p_T) - \log(1-p_T)   
		\\&- \sum_sv_s\left(\log\left(\frac{\fpr_s}{1-\fpr_s}\right) - \log\left(\frac{\tpr_s}{1-\tpr_s}\right)\right)
		\\&- \sum_s\left(\log(1-\fpr_s) -\log(1-\tpr_s)\right)\\
		\stackrel{(\ref{eqn:resolve_wi})}={}& \log(p_T) - \log(1-p_T)   + \sum_sv_sw_s 
		\\&+ \sum_s\left(\log(1-\tpr_s) - \log(1-\fpr_s)\right)\\
	\Rightarrow 
	b ={}& \log\left(\frac{p_T}{1-p_T}\right) + \sum_s\left(\log(1-\tpr_s) - \log(1-\fpr_s)\right)
\end{align*}


\end{document}